\documentclass[10pt,twocolumn,letterpaper]{article}

\usepackage[pagenumbers]{iccv} 

\definecolor{iccvblue}{rgb}{0.21,0.49,0.74}
\usepackage[pagebackref,breaklinks,colorlinks,allcolors=iccvblue]{hyperref}
\usepackage[dvipsnames]{xcolor}
\usepackage{multirow}
\usepackage{graphicx}
\usepackage{booktabs}
\usepackage{enumitem}
\usepackage{multirow}
\usepackage{subcaption}
\usepackage{fontawesome5}
\usepackage[subtle]{savetrees}
\usepackage{pifont}
\usepackage{wrapfig}
\usepackage{adjustbox}
\usepackage{tabularx} 
\usepackage[normalem]{ulem}
\usepackage{colortbl}    
\newcommand{\cmark}{\ding{51}}%
\newcommand{\xmark}{\ding{55}}%

\def\paperID{6475} 
\def\confName{ICCV}
\def\confYear{2025}

\title{SuperEdit: Rectifying and Facilitating \uline{Super}vision for Instruction-Based Image \uline{Edit}ing\\ 
  \vspace{0.5em}  
  \large 
  \hspace{2pt}\href{https://liming-ai.github.io/SuperEdit/}{\faGlobe\hspace{2pt}Website} \quad 
  \hspace{2pt}\href{https://github.com/bytedance/SuperEdit}{\faGithub\hspace{2pt}Code} \quad \hspace{2pt}\href{https://huggingface.co/datasets/limingcv/SuperEdit-40K}{\raisebox{-0.3\height}{\includegraphics[height=1.4em]{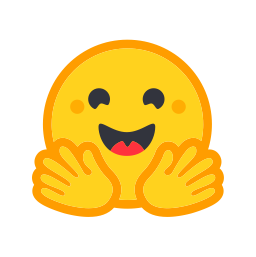}}\hspace{2pt}Data}\vspace{-0.8cm}}

\author{Ming Li$^{1,2,\ddagger}$, Xin Gu$^{1}$, Fan Chen$^{1}$, Xiaoying Xing$^{1}$, Longyin Wen$^{1}$, Chen Chen$^{2}$, Sijie Zhu$^{1,*}$ \\
\scalebox{0.9}{$^{1}$ByteDance Intelligent Creation (USA) \quad $^{2}$Center for Research in Computer Vision, University of Central Florida}
}

\begin{document}
\twocolumn[{%
\renewcommand\twocolumn[1][]{#1}%
\maketitle
\begin{center}
    \vspace{-20pt}
    \includegraphics[width=1.0\linewidth]{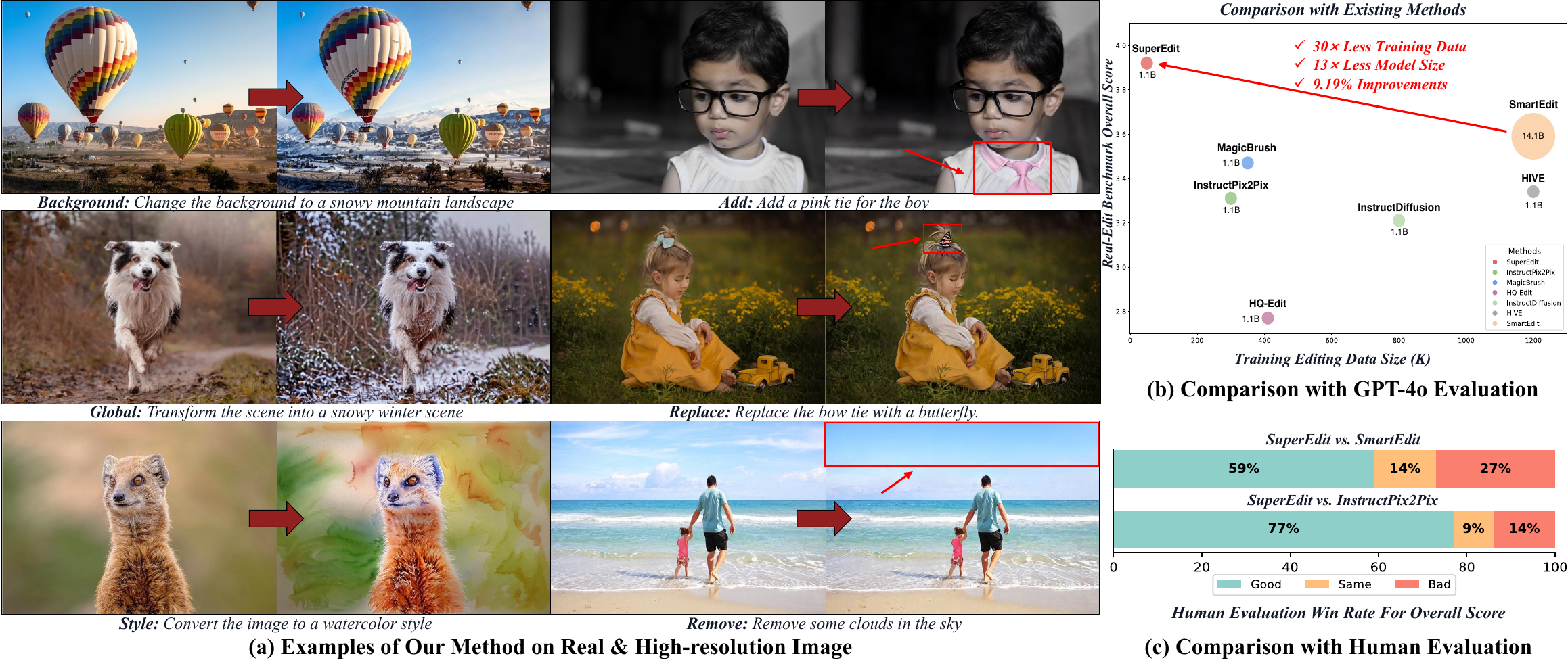}
\vspace{-0.7cm}
\captionof{figure}{
\textbf{(a)} Our editing method works well with real and high-resolution images, handling various free-form edits (left) and local edits (right); \textbf{(b)} Compared to the current state-of-the-art SmartEdit, our method achieves a 9.19\% performance improvement with 30$\times$ less training data and 13$\times$ fewer model parameters; \textbf{(c)} Our method achieves better overall scores on the human evaluation results, indicating more precise editing capabilities.
}
    \vspace{3pt}
    \label{fig:teaser}
    \end{center}%
}]
\let\thefootnote\relax\footnotetext{$*$ Corresponding author, sijiezhu@bytedance.com}
\let\thefootnote\relax\footnotetext{$\ddagger$ This work was done during the internship at ByteDance, San Jose, USA}
\begin{abstract}
Due to the challenges of manually collecting accurate editing data, existing datasets are typically constructed using various automated methods, leading to noisy supervision signals caused by the mismatch between editing instructions and original-edited image pairs. Recent efforts attempt to improve editing models through generating higher-quality edited images, pre-training on recognition tasks, or introducing vision-language models (VLMs) but fail to resolve this fundamental issue. In this paper, we offer a novel solution by constructing more effective editing instructions for given image pairs. This includes rectifying the editing instructions to better align with the original-edited image pairs and using contrastive editing instructions to further enhance their effectiveness. Specifically, we find that editing models exhibit specific generation attributes at different inference steps, independent of the text. Based on these prior attributes, we define a unified guide for VLMs to rectify editing instructions. However, there are some challenging editing scenarios that cannot be resolved solely with rectified instructions. To this end, we further construct contrastive supervision signals with positive and negative instructions and introduce them into the model training using triplet loss, thereby further facilitating supervision effectiveness. Our method does not require the VLM modules or pre-training tasks used in previous work, offering a more direct and efficient way to provide better supervision signals, and providing a novel, simple, and effective solution for instruction-based image editing. Results on multiple benchmarks demonstrate that our method significantly outperforms existing approaches. Compared with previous SOTA SmartEdit, we achieve 9.19\% improvements on the Real-Edit benchmark with 30$\times$ less training data and 13$\times$ smaller model size. All data and models are open-sourced on \href{https://github.com/bytedance/SuperEdit}{Github} for future research.
\end{abstract}
\vspace{-0.5cm}
\section{Introduction}
In recent years, significant progress has been made in text-to-image (T2I) generation~\cite{sd,sdxl,sd3,dalle,dalle2} due to the development of diffusion models~\cite{diffusion,diffusion_beat_gan,ddpm,ddim}. These T2I diffusion models can generate images that align with natural language descriptions while satisfying human perception and preferences. Consequently, numerous image editing methods~\cite{sdedit,p2p,instruct_p2p,diffedit} based on these models have been proposed to achieve various editing effects. Instruction-based methods~\cite{instruct_p2p,smartedit,mgie} have become increasingly popular as they allow users to conveniently and easily modify images using language instructions without the need to provide masks, as required by mask-based methods~\cite{smartbrush,learnable_region,smartmask,foi}.

The training of instruction-based editing models requires the original-edited image pairs and corresponding editing instruction, making it difficult to manually create or collect a large amount of relevant data~\cite{magicbrush}. To address the issue of scarce training data, existing efforts~\cite{p2p,hq_edit,ultraedit,seed_data_edit} have attempted to develop various automated pipelines to synthesize large datasets. Specifically, most methods first use large language models (LLMs) to modify the text descriptions of original images. The original images and modified texts are then input into various pre-trained diffusion models to automatically generate edited images. However, current text-to-image diffusion models struggle to fully correspond to input text prompts~\cite{geneval, controlnet}. This often changes parts of the original images that do not require editing, leading to misaligned editing instructions and original-edited image pairs, thus resulting in noisy supervision signals. To mitigate the potential issues of noisy supervision in image editing models, existing work has attempted to introduce additional recognition pre-training tasks for U-Net~\cite{unet} such as semantic segmentation~\cite{instructdiffusion,emu_edit}, or replace CLIP~\cite{clip} text encoder with vision-language models (VLMs)~\cite{smartedit,mgie} to better understand editing instructions from noisy supervision signals. However, these methods not only introduce significant computational overhead but also overlook the issue of noisy supervision signals.

\begin{figure}[t!]\centering
    \includegraphics[width=0.9\linewidth]{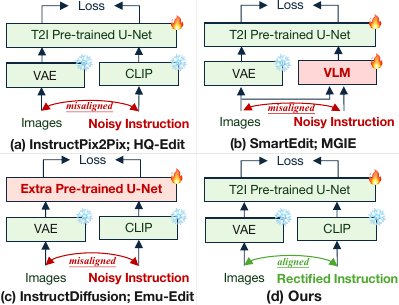}
    \vspace{-0.3cm}
    \caption{
    Unlike existing efforts that attempt to \textbf{(a)} scale up edited images with noisy supervision~\cite{instruct_p2p,hq_edit}, \textbf{(b)} introduce massive VLMs into editing model architecture~\cite{smartedit,mgie}, and \textbf{(c)} perform additional pre-training tasks~\cite{instructdiffusion,emu_edit}, \textbf{(d)} we focus on improving the effectiveness of supervision signals, which is the fundamental issue of image editing. 
    }
    \vspace{-0.6cm}
    \label{fig:motivation}
\end{figure}

In this paper, we focus on addressing the fundamental challenge by introducing more effective editing instructions, as demonstrated in Fig.~\ref{fig:motivation}. Our data-oriented method explores a different research question: how much performance improvement can be achieved solely by focusing on supervision signal quality and optimization in image editing? Surprisingly, SuperEdit outperforms existing methods in both GPT-4o and human evaluations, despite using less data and requiring no additional modules or pretraining as shown in Fig.\ref{fig:teaser}. This demonstrates that high-quality supervision signals can significantly compensate for architectural simplicity, achieving results comparable to or better than methods with more complex requirements. 

\begin{figure*}[ht!]\centering
    \includegraphics[width=1.0\linewidth]{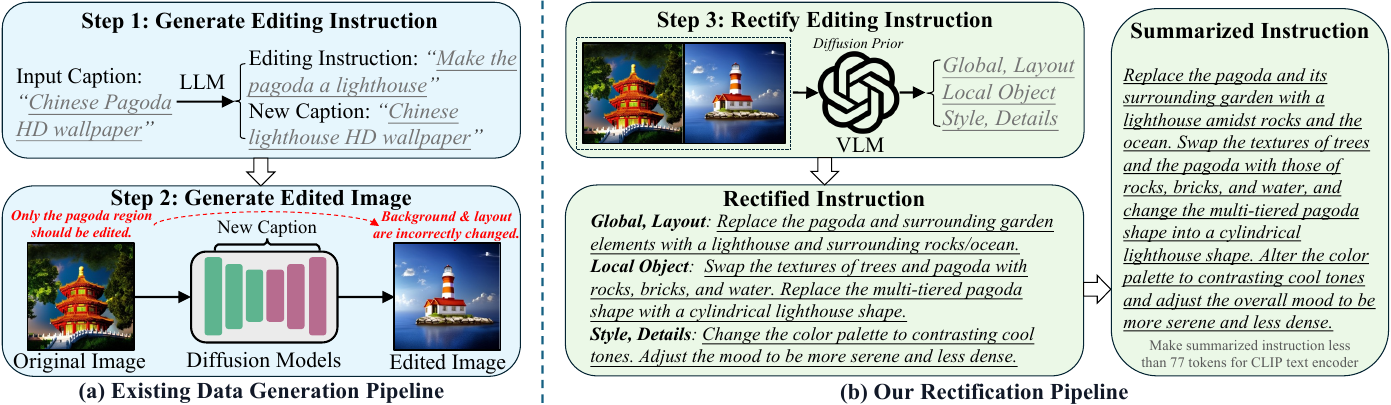}
    \vspace{-0.6cm}
    \caption{
    \textbf{(a)} Existing work primarily uses LLMs and diffusion models to automatically generate edited images. However, current diffusion models often fail to accurately follow text prompts while maintaining the input image's layout, resulting in mismatches between the original-edited image pairs and the editing instructions. \textbf{(b)} We perform instruction rectification (Step 3) based on the images constructed in Steps 1 and 2. We show VLMs can understand the differences between the images, enabling them to rectify editing instructions to be better aligned with original-edited image pairs.
    }
    \label{fig:data}
\end{figure*}

Specifically, to enhance the effectiveness of supervision signals for instruction-based image editing methods, we propose using VLMs to rectify editing instructions, creating better-aligned instructions for the original-edited image pairs. However, determining which VLM to use for this task and how to establish a unified rectification method for various editing instructions remain unexplored problems. To address this, we first analyze the ability of different VLMs to understand the differences between original and edited images, showing that GPT-4o~\cite{gpt4} is the most capable of rectifying editing instructions.  Additionally, we observe that both editing models and text-to-image diffusion models share a similar prior, as shown in Fig.~\ref{fig:diffusion_prior}: different inference stages correspond to the generation of different image attributes, independent of the input text prompt~\cite{p2p,tgate,ediff,localizing_diffusion,understanding_diffusion,focus_instruction,lime} or editing instructions. Inspired by this, we guide VLMs based on these attributes to establish a unified rectification guideline for various editing instructions, as demonstrated in Fig.~\ref{fig:data}.

When training with only rectified editing instructions, we find that the editing model can better understand the editing commands but still faces challenges in handling complex scenarios. For example, when the original image contains multiple objects, the edit model struggles to perform an accurate editing function if the instructions modify only one of these objects. Additionally, inherent issues present in pre-trained text-to-image diffusion models~\cite{ella,geneval,T2i-compbench,smartedit}, such as difficulty in understanding quantity, position, or object relationships, persist in the editing models. To address these issues, we propose using contrastive supervision signals to further optimize the editing model. Specifically, we first construct incorrect editing instructions based on the rectified instructions to generate positive and negative samples. We then introduce a triplet loss to guide the model, thereby enhancing the effectiveness of supervision, as shown in Fig.~\ref{fig:pipeline}.

In summary, our contributions are summarized as follows:
\begin{itemize}[leftmargin=*]
\item \underline{\textit{New Insight}}: 
We aim to address the noisy supervision problem that arises from the misalignment between editing instructions and original-edited image pairs, which is a fundamental issue overlooked by previous work, as shown in Fig.~\ref{fig:motivation}.

\item \underline{\textit{Rectifying Supervision}}: 
We leverage diffusion generation priors to guide the vision-language model to generate better-aligned editing instructions for original-edited image pairs.

\item \underline{\textit{Facilitating Supervision}}: 
We introduce contrastive supervision using triplet loss, enabling the editing model to learn from both positive and negative editing instructions.

\item \underline{\textit{Promising Results}}:
We achieve significant improvements on multiple benchmarks without additional pre-training or VLM. Compared to SmartEdit~\cite{smartedit}, we achieved a 9.19\% improvement while reducing 30$\times$ data and 13$\times$ model parameters.
\end{itemize}

\section{Related Work}
\label{sec:related}

\subsection{Image Editing with Diffusion Models}
Building on advancements in text-to-image (T2I) diffusion models~\cite{sd, sdxl, sd3, dalle, dalle2, imagen}, recent research has explored them for image editing~\cite{p2p, instruct_p2p}. Training-free methods~\cite{p2p, pnp, sdedit, masactrl, zero_i2i, attn_edit} typically achieve this by adjusting attentions in pre-trained T2I models, but have limited performance and generalization capabilities on various editing tasks.

Training-based methods address these limitations with specialized editing models, which can be categorized into mask-based and instruction-based approaches. Mask-based methods~\cite{inpaint_anything,diffedit,learnable_region,foi,smartbrush,smartmask} enable fine-grained local edits with user-provided or predicted masks and corresponding text descriptions. However, it struggles with global image editing and is constrained by the lack of mask-based editing data~\cite{smartedit}.

Instruction-based methods directly accept textual commands, such as ``add a dog", offering better editing flexibility and generalization. InstructPix2Pix~\cite{instruct_p2p} pioneered this paradigm by generating instruction-based editing data and modifying the conditions of T2I diffusion models. Building on this framework, subsequent work introduces vision-language models~\cite{Kosmos-G,mgie,smartedit} or additional pre-training tasks for the denoising U-Net~\cite{instructdiffusion,smartedit,emu_edit,unet} to enhance the understanding and reasoning of input conditions. \textit{However, these methods not only introduce substantial computational overhead but also overlook the fundamental noisy supervision issue.}

\subsection{Generating and Improving Editing Supervision}
Due to the difficulty of scaling instruction-based image editing data through manual collection, existing efforts~\cite{instruct_p2p,magicbrush,seed_data_edit,ultraedit,hq_edit} aim to automatically modify text descriptions of original images and generate edited images with T2I diffusion models. However, this approach often produces synthesized images that do not align with the editing instructions, as shown in Figure~\ref{fig:data}, resulting in noisy editing supervision signals~\cite{magicbrush,ultraedit}. To address this, MagicBrush~\cite{magicbrush} manually filters out incorrect editing data, but it is hard to scale. Unlike existing methods focusing on edited image quality, we leverage diffusion prior and vision-language model (i.e., GPT-4o~\cite{gpt4}) to create better-aligned instructions with original-edited image pairs, providing more accurate supervision.

\subsection{Alignment of Diffusion Models}
The success of alignment training in large language models (LLMs)~\cite{instruct_gpt,dpo,rlaif} has been applied to diffusion models for better image generation. This is achieved by maximizing reward scores~\cite{imagereward,cpp,prdp} or the generation probability of the winner image in a pair~\cite{diffusion_dpo,dpok,D3PO}. In image editing, HIVE~\cite{hive} and MultiReward~\cite{multireward} attempt to incorporate reward information into the text condition to align the editing model. In contrast, we guide the editing model by rectifying and constructing contrastive editing instructions, achieving more effective alignment.


\section{Method}
\label{sec:method}

In this section, we first introduce the most general image editing framework in Sec.~\ref{subsec:editing_framework}. Then, we explain how to use diffusion priors to rectify editing instructions with the multimodal model (i.e., GPT-4o) in Sec.~\ref{subsec:rectifying}, thereby enhancing the accuracy of supervision signals. Finally, we describe how to construct contrastive supervision with both correct and incorrect editing instructions and integrate it into the editing model training using triplet loss in Sec.~\ref{subsec:facilitating}.

\subsection{Instruction-based Image Editing Framework}
\label{subsec:editing_framework}
InstructPix2Pix~\cite{instruct_p2p} pioneered instruction-based image editing, performing editing tasks by simultaneously taking the original image $C^{I}$ and editing instructions $C^{T}$ as input conditions to generate the edited image $x$ from random noise $\epsilon$. Following the definition of DDPM~\cite{ddpm}, we randomly sample a timestep $t \in T$ during training, and then add corresponding noise $\epsilon_t$ to the edited image $x$:
\vspace{-0.05cm}
\begin{equation}
x_t=\sqrt{\bar{\alpha}_t} x+\sqrt{1-\bar{\alpha}_t} \epsilon_t, \quad \epsilon \sim \mathcal{N}(\mathbf{0}, {I}),
\label{eq:add_noise}
\end{equation}
\vspace{-0.05cm}
where $\epsilon$ is a noise map sampled from a Gaussian distribution, and $\bar{\alpha}_t:=\prod_{s=0}^t \alpha_s$, $\alpha_t = 1 - \beta_t$ is a differentiable function of timestep $t$, which is determined by the denoising sampler such as DDPM~\cite{ddpm}. Then the training objective of the editing model $\epsilon_\theta$ is predicting the added noise at timestep $t$, which can be written as:
\vspace{-0.05cm}
\begin{equation}
\resizebox{0.9\linewidth}{!}{%
$
    \mathcal{L}_{\text {train }}=\mathbb{E}_{x_{t}, t, C^{I}, C^{T}, \epsilon}\left[\left\|\epsilon_\theta\left(\text{concat}(x_t, C^{I}), t, C^{T}\right)-\epsilon_t\right\|_2^2\right],
$
}
\end{equation}
\vspace{-0.05cm}
where $\text{concat}$ refers to concatenating the image latents of noised edited image $x_{t}$ and original image $c_I$ in the channel dimension.

\subsection{Rectifying Supervision with Diffusion Priors}
\label{subsec:rectifying}

\begin{figure}[t!]\centering
    \includegraphics[width=1.0\linewidth]{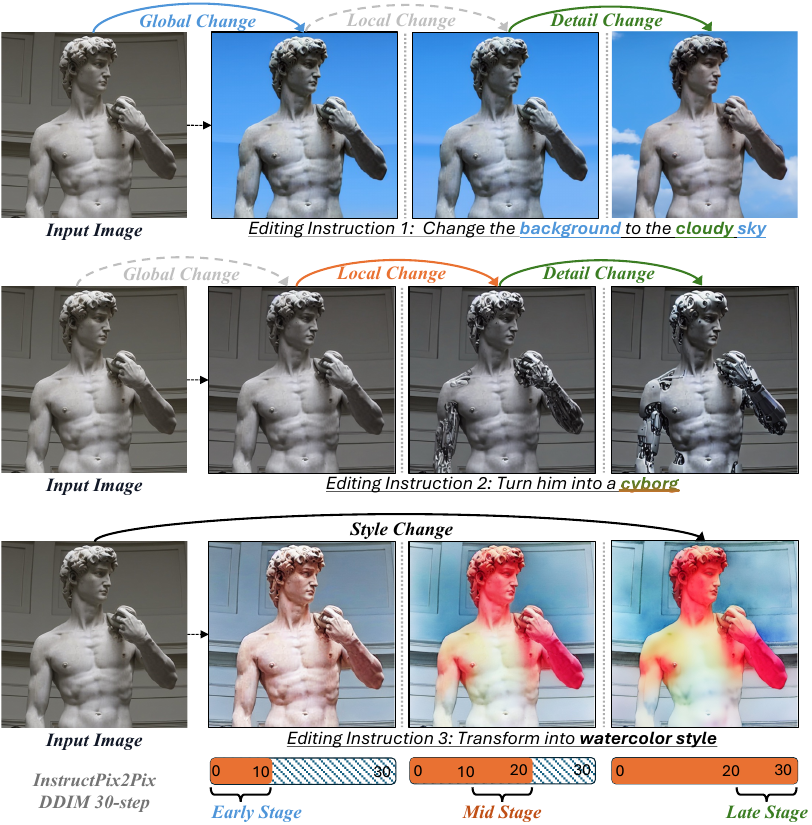}
    \vspace{-0.7cm}
    \caption{
    We show that the editing model follows consistent generation attributes at different sampling stages, independent of the editing instructions. The early, middle, and late sampling stages correspond to \textbf{global}, \textbf{local}, and \textbf{detail} changes, respectively, while \textbf{style} changes occur at all stages. All the generated images here are DDIM 30-step sampled final images. The \textcolor[HTML]{e97132}{orange progress bar} and the \textcolor[HTML]{6597ad}{grid progress bar} represent the sampling stages with and without the editing instructions, respectively. 
    }
    \vspace{-0.4cm}
    \label{fig:diffusion_prior}
\end{figure}

As shown in Fig.~\ref{fig:data}, existing image editing datasets~\cite{instruct_p2p,magicbrush,seed_data_edit} typically use only Steps 1 and 2: LLMs construct editing prompts and captions, and then text-to-image diffusion models synthesize edited images. However, diffusion models often fail to accurately follow prompts while maintaining image layout, creating mismatches between original-edited pairs and editing instructions, resulting in inaccurate supervision. While better supervision signals for text-to-image diffusion models are common in image generation~\cite{dalle3,kolors}, this approach remains underexplored in image editing due to two challenges: (1) VLMs trained on single-image data struggle with multi-image inputs, and (2) editing instructions vary widely, making unified rectification guidelines difficult. To address these issues, we: (1) analyzed different VLMs' capabilities with multi-image inputs, finding GPT-4o most effective, and (2) discovered that timestep-specific roles in image generation also apply to editing, providing a foundation for a unified rectification method across various instructions (Fig.~\ref{fig:data} and \ref{fig:diffusion_prior}). Due to page limitations, our VLM analysis is in the \textcolor{blue}{Supplementary Material}, while this section focuses on Diffusion Prior and Editing Instruction Rectification.

\begin{figure*}[ht!]\centering
    \includegraphics[width=1.0\linewidth]{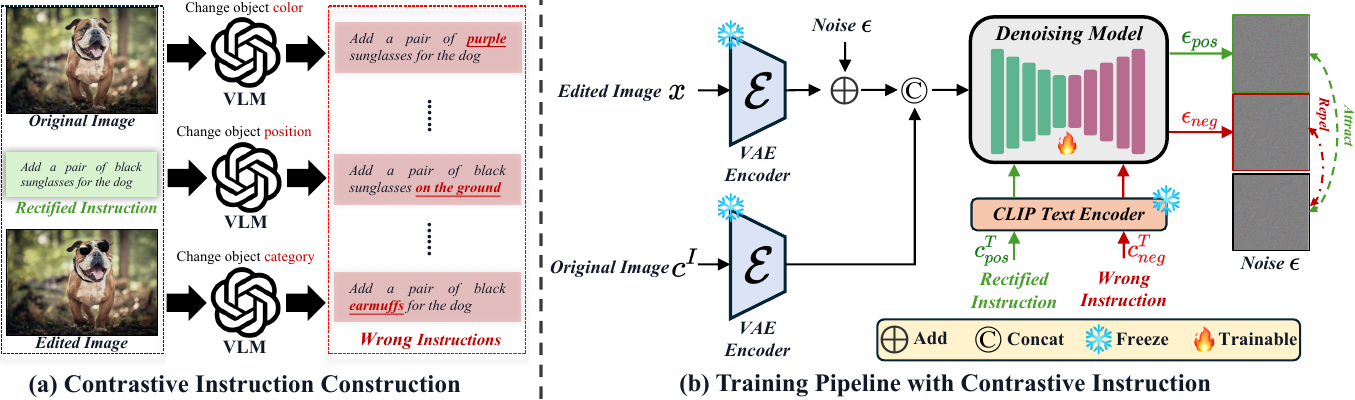}
    \vspace{-0.5cm}
    \caption{
    \textbf{(a)} Based on the rectified editing instruction and original-edited image pair, we utilize the Vision-Language Models (VLM) to generate various image-related wrong instructions. These involve random substitutions of quantities, spatial locations, and objects within the rectified editing instructions according to the original-edited images context; \textbf{(b)} During each training iteration, we randomly select one wrong instruction $\textcolor{black}{c^T_{neg}}$ and input it along with the rectified instruction $\textcolor{black}{c^T_{pos}}$ into the editing model to obtain predicted noises. The goal is to make the rectified instruction’s predicted noise $\textcolor{black}{\textbf{$\epsilon_{pos}$}}$ closer to the sampled training diffusion noise $ \epsilon$, while ensuring the noise from incorrect instructions $\textcolor{black}{\textbf{$\epsilon_{neg}$}}$ is farther. Best viewed in color.
    }
    \vspace{-0.2cm}
    \label{fig:pipeline}
\end{figure*}

\vspace{-0.3cm}
\paragraph{Diffusion Generation Priors.}
\label{diffusion_prior}
Previous work has shown that different timesteps play distinct roles in image generation for text-to-image diffusion models, regardless of the text prompt~\cite{p2p,tgate,ediff,localizing_diffusion,understanding_diffusion,focus_instruction,lime}. We find that this phenomenon also exists in instruction-based editing models and present examples based on pre-trained InstructPix2Pix~\cite{instruct_p2p}, as shown in Fig.~\ref{fig:diffusion_prior}. Specifically, diffusion models focus on global layout in the early stages, local object attributes in the mid stages, and image details in the late stages of sampling.  This finding inspires us to guide VLMs based on these four generation attributes, establishing a unified rectification method for various editing instructions. We provide more analysis and details in the \textcolor{blue}{Supplementary Material}.

\vspace{-0.3cm}
\paragraph{Editing Instruction Rectification.}
\label{instruction_rectification}
As demonstrated in Fig.~\ref{fig:data}, we extend the existing editing data generation pipeline by introducing our instruction rectification (Step 3). This process relies on the original edited image pairs obtained through Steps 1 and 2 from previous work. Specifically, we input original-edited image pairs into the vision-language model (i.e., GPT-4o) and instruct it to describe the changes in the edited image compared to the original image according to the above diffusion prior generation attributes. Finally, we use VLM to summarize the instructions and ensure that its length is less than the maximum length of CLIP text encoder, which is 77 tokens.

\vspace{-0.05cm}
\subsection{Facilitating Supervision with Contrastive Instructions}
\vspace{-0.05cm}
\label{subsec:facilitating}
Although using rectified editing instructions can significantly improve performance across various editing tasks, we find that editing models still struggle with closely related text instructions. For example, \textit{``add a cat on the left side of the image"} and \textit{``add two cats on the right side of the image"} might produce the same edited image. This indicates that inherent biases in pre-trained text-to-image diffusion models~\cite{geneval,ella}, such as difficulties in understanding quantity, position, and spatial relationships, persist in editing models. More importantly, our experiments show that training models with rectified editing instructions does not resolve these challenges. To further facilitate supervision signal effectiveness, we drew on successful alignment experiences from large language models~\cite{dpo,gpt4,instruct_gpt} and text-to-image diffusion models~\cite{imagereward,diffusion_dpo,noise_clr}: constructing positive and negative sample pairs and guiding the model to assign a higher generation probability to positive samples compared to negative ones.

\vspace{-0.2cm}
\paragraph{Constructing Contrastive Instructions.} Unlike the standard alignment process for large language models or text-to-image diffusion models, it is challenging to generate different editing results from the same instruction to create positive and negative sample pairs for image editing tasks. To address this, we construct positive and negative editing instructions for alignment, thereby generating relatively positive and negative edited images. As shown in Fig.~\ref{fig:pipeline} (a), we use the original image, edited image, and rectified editing instruction as input. The VLM (i.e., GPT-4o) is used to modify attributes in the rectified editing instruction, such as quantity, spatial relationships, and object types, to create different wrong instructions. Here, we require VLM to modify only a single attribute from the rectified editing instruction in each wrong instruction, keeping most of the editing text unchanged. Since only a few words are replaced between the rectified instruction and the wrong instruction, the text embeddings produced by the CLIP text encoder that serve as input to the denoising model will also be similar. This ensures the task's learning difficulty, helping the model understand how subtle differences between the two editing instructions result in significantly different editing results. 

\vspace{-0.4cm}
\paragraph{Facilitating Editing Models with Contrastive Instructions.}
Our key insight is that enhancing the effectiveness of supervision signals can improve various editing tasks without introducing additional model architectures or pre-training tasks. Therefore, we adhere strictly to the InstructPix2Pix~\cite{instruct_p2p} model architecture and training pipeline. To be specific, the inputs including the original image $c^{I}$, edited image $x$, the rectified instruction $c^T_{pos}$, and wrong editing instruction $c^T_{neg}$. During training, we will add a sampled timestep $t \in T$ to obtain the noised edited image $x_t$ with Equation~\ref{eq:add_noise}. Both the rectified and wrong editing instructions are fed into the denoising model to predict the final noises $\epsilon_{pos}$ and $\epsilon_{neg}$, which are then used to construct positive and negative samples, respectively:
\begin{align}
\epsilon_{pos} &= \epsilon_\theta\left(\operatorname{concat}\left(x_t,\, c^I\right),\, t,\, c^T_{pos}\right), \\
\epsilon_{neg} &= \epsilon_\theta\left(\operatorname{concat}\left(x_t,\, c^I\right),\, t,\, c^T_{neg}\right).
\end{align}

After constructing the positive and negative sample pairs, we aim for the noise predicted by the positive editing instruction $\epsilon_{pos}$ to be closer to the true noise $\epsilon_{t}$ sampled during training, compared to the noise predicted by the wrong editing instruction $\epsilon_{neg}$. This goal can be achieved through a triplet loss function:
\begin{equation}
    \mathcal{L}_{\text{triplet}} = max\{d(\epsilon_{t}, \epsilon_{pos}) - d(\epsilon_{t}, \epsilon_{neg}) + m, 0\},
\end{equation}
where $d\left(x, y\right)=\left\|\mathbf{x}-\mathbf{y}\right\|_2^2$ and margin $m$ is a hyper-parameter. The final training loss is the combination of the original diffusion training loss and the triplet loss:
\begin{align}
    \mathcal{L}_{\text{total}} = \mathcal{L}_{\text{train}} + \lambda \cdot \mathcal{L}_{\text{triplet}}, \text{ where }\mathcal{L}_{\text{train}} = d(\epsilon_{t}, \epsilon_{pos}).
\end{align}

Please note that the contrastive supervision signals are only used during the training phase. During inference, the editing model only requires one single input editing instruction.

\section{Experiment}
\label{sec:experiment}

\begin{figure*}[h!]\centering
    \includegraphics[width=0.94\linewidth]{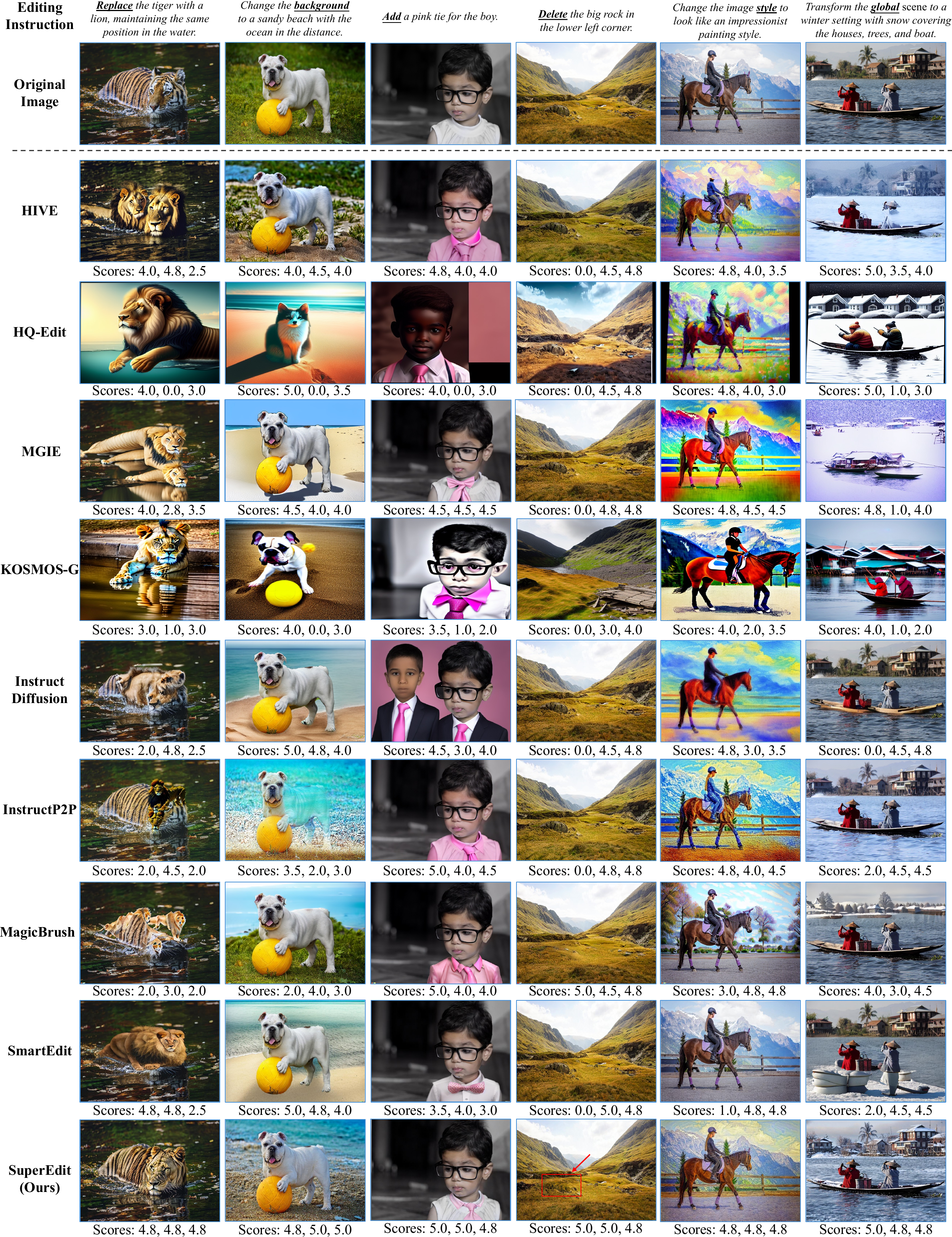}
    \vspace{-0.3cm}
    \caption{
    Visual comparison with existing methods and the corresponding human-aligned GPT-4o evaluation scores (Following, Preserving, Quality Scores from left to right). We achieve better results while preserving the layout, quality, and details of the original image. Please note that we do not claim that our editing results are flawless. We provide more visual comparison results in the supplementary material.
    }
    \label{fig:visual_comparison}
\end{figure*}

\begin{table*}[h]
\centering
\vspace{-0.2cm}
\resizebox{0.9\linewidth}{!}{
\begin{tabular}{c|cccc|cccccccc}
\specialrule{1.5pt}{0pt}{0pt}
\multirow{2}{*}{Method} & \multirow{2}{*}{\begin{tabular}[c]{@{}c@{}}Extra\\ Module\end{tabular}} & \multirow{2}{*}{\begin{tabular}[c]{@{}c@{}}Pretrain\\ Tasks\end{tabular}} & \multirow{2}{*}{\begin{tabular}[c]{@{}c@{}}Edit\\ Data\end{tabular}} & \multirow{2}{*}{\begin{tabular}[c]{@{}c@{}}Model\\ Size\end{tabular}} & \multicolumn{2}{c}{Following $\uparrow$} & \multicolumn{2}{c}{Preserving $\uparrow$} & \multicolumn{2}{c}{Quality $\uparrow$} & \multicolumn{2}{c}{Overall $\uparrow$} \\ 
& & & & &  Acc & Score & Acc & Score & Acc & Score & Acc & Score \\ \hline
KOSMOS-G~\cite{Kosmos-G} & \cmark & \cmark & 9.0M & 1.9B & 51\% & 2.82 & 9\% & 1.43 & 27\% & 3.20 & 29.0\% & 2.48 \\
MGIE~\cite{mgie} & \cmark & \cmark & 1.0M & 8.1B & 40\% & 2.43 & 45\% & 2.79 & 38\% & 3.35 & 41.0\% & 2.86 \\
SmartEdit~\cite{smartedit} & \cmark & \cmark & 1.2M & 14.1B & 64\% & 3.50 & 66\% & 3.70 & 45\% & 3.56 & 58.3\% & 3.59 \\
MultiReward~\cite{multireward} & \cmark & \cmark & 320K & 1.2B & 63\% & 3.39 & 58\% & 3.43 & 54\% & 3.80 & 58.3\% & 3.54 \\ 
InstructDiffusion~\cite{instructdiffusion} & \xmark & \cmark & 860K & 1.1B & 52\% & 2.87 & 54\% & 3.17 & 45\% & 3.58 & 50.3\% & 3.21 \\
InstructPix2Pix~\cite{instruct_p2p} & \xmark & \xmark & 300K & 1.1B & 52\% & 2.94 & 53\% & 3.31 & 50\% & 3.69 & 51.7\% & 3.31 \\
MagicBrush~\cite{magicbrush} & \xmark & \xmark & 310K & 1.1B & 51\% & 2.90 & 70\% & 3.85 & 50\% & 3.67 & 57.0\% & 3.47 \\
HIVE~\cite{hive} & \xmark & \xmark & 1.1M & 1.1B & 54\% & 2.93 & 56\% & 3.36 & 53\% & 3.72 & 54.3\% & 3.34 \\
HQ-Edit~\cite{hq_edit} & \xmark & \xmark & 500K & 1.1B & 51\% & 2.84 & 16\% & 1.63 & 54\% & 3.84 & 40.3\% & 2.77 \\
SuperEdit (Ours) & \xmark & \xmark & \textbf{40K} & 1.1B & \textbf{67\%} & \textbf{3.59} & \textbf{77\%} & \textbf{4.14} & \textbf{65\%} & \textbf{4.01} & \textbf{69.7\%} & \textbf{3.91} 
\\ \specialrule{1.5pt}{0pt}{0pt}
\end{tabular}
}
\vspace{-0.2cm}
\caption{Comparison with instruction-based image editing methods on Real-Edit benchmark~\cite{multireward}. Compared to existing work, our method achieves state-of-the-art performance across all metrics using a small amount of high-quality editing data without introducing additional models or pre-training tasks. Please note that the scores range from 0 to 5. $\uparrow$ denotes a higher result is better. All baseline results are cited from the MultiReward~\cite{multireward} paper.}
\label{tab:real_edit_benchmark}
\vspace{-0.1cm}
\end{table*}

\begin{figure*}[ht!]\centering
    \includegraphics[width=1.0\linewidth]{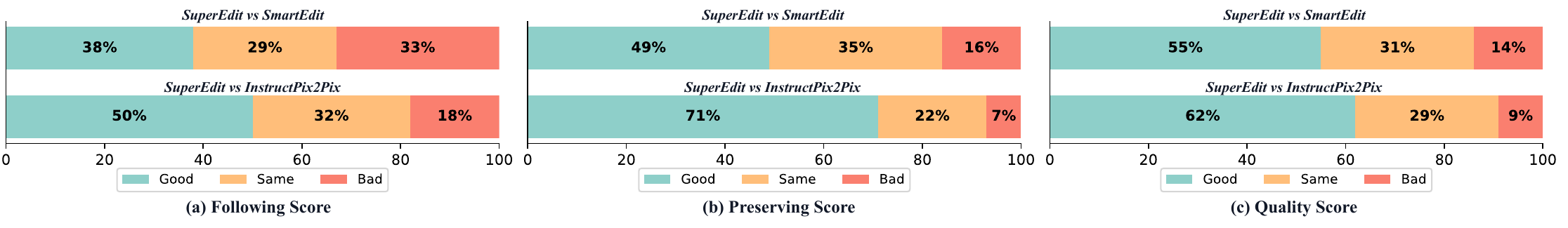}
    \vspace{-0.8cm}
    \caption{
    Human evaluation on three evaluation criteria for image editing effects. \textbf{(a)} Following: whether the edited image adhere to the editing instructions; \textbf{(b)} Preserving: whether the image structure outside of the editing instructions has been preserved;  \textbf{(c)} Quality: whether the overall quality/aesthetics of the edited image has been degraded compared to the input image. Our SuperEdit achieves the best results on all of these metrics.
    }
    \vspace{-0.2cm}
    \label{fig:human_eval}
\end{figure*}

\begin{table}[th!]
\resizebox{\linewidth}{!}{
\begin{tabular}{lllll}
\specialrule{1.5pt}{0pt}{0pt}
 & Following $\uparrow$ & Preserving $\uparrow$ & Quality $\uparrow$ & Overall $\uparrow$ \\
 \hline
InstructPix2Pix~\cite{instruct_p2p} & 2.41 & 2.62 & 2.44 & 2.49 \\
SmartEdit-13B~\cite{smartedit} & 3.09 & 3.06 & 2.63 & 2.93 \\
SuperEdit & \textbf{3.18\textsubscript{\textcolor{red}{\textbf{+1.80\%}}}} & \textbf{3.86\textsubscript{\textcolor{red}{\textbf{+16.00\%}}}} & \textbf{3.37\textsubscript{\textcolor{red}{\textbf{+14.80\%}}}} & \textbf{3.47\textsubscript{\textcolor{red}{\textbf{+10.80\%}}}}
\\ \specialrule{1.5pt}{0pt}{0pt}
\end{tabular}
}
\vspace{-0.3cm}
\caption{Human evaluation results on Real-Edit~\cite{multireward} benchmark. All the human-evaluated scores range from 0 to 5. Overall represents the average score of Following, Preserving, and Quality scores.}
\label{tab:human_eval}
\vspace{-0.5cm}
\end{table}

\subsection{Data Collection and Construction}
To build a diverse dataset with various types of editing instructions, we need original and edited images from different data domains, as well as a wide variety of editing instructions. To achieve this, we sampled data from different public editing datasets to construct rectified and contrastive supervision signals. Specifically, we extracted 10,177, 8,807, and 21,016 editing pairs from InstructPix2Pix~\cite{instruct_p2p}, MagicBrush~\cite{magicbrush}, and Seed-Data-Edit~\cite{seed_data_edit}, respectively, resulting in a total of 40,000 training samples. During extraction, we strive to ensure that the data for different types of editing tasks is as balanced as possible. We then applied our proposed methods in Sec.~\ref{sec:method} to rectify and construct contrastive editing instructions for these training samples. Since the MagicBrush data has been manually verified, we skip the rectification step for this dataset and directly construct contrastive supervision based on the original editing instructions. For Seed-Data-Edit dataset, we only sample images from the first part of data without human editing instructions.

\subsection{Experimental Settings}

\paragraph{Evaluation Benchmarks and Metrics.}
To more accurately assess the effectiveness of various editing models, we conducted assessments on the Real-Edit benchmark~\cite{multireward}, which is a human-aligned evaluation benchmark with GPT-4o scoring. Specifically, MultiReward~\cite{multireward} uses high-resolution images from the Unsplash community as a test dataset and combines them with GPT-4o~\cite{gpt4} to create an automated evaluation method for single-turn editing. It assesses edited images in terms of accuracy (\%) and scores (from 0 to 5), evaluating whether they adhere to the editing instructions (Following), whether the image structure outside of the editing instructions has been preserved (Preserving), and whether the overall quality/aesthetics of the edited image has been degraded compared to the original one (Quality).

\subsection{Experimental Results}
\paragraph{Comparison on Real-Edit Benchmark.}
In Tab.~\ref{tab:real_edit_benchmark}, we present the quantitative results of editing effectiveness on the Real-Edit benchmark~\cite{multireward}. Without introducing additional parameters or pre-training stages, our method achieves the best results in the three GPT-4o automated evaluation metrics: Following, Preserving, and Quality, each of which includes percentage accuracy (Acc) and scores (from 0 to 5). For example, compared to SmartEdit~\cite{smartedit}, which introduces an additional 13B vision-language model (i.e., LLaVA~\cite{llava}) to the 1.1B InstructPix2Pix~\cite{instruct_p2p} framework, we achieved improvements of 11.4\% Overall Score. This suggests that given accurate and effective supervision signals, the trained editing model can understand and successfully execute the editing instructions, without the need for additional vision-language models.

It is worth noting that unlike existing image editing methods, which often show improvement in a single metric while others remain unchanged or worsen, our method achieves comprehensive and significant advancements across all three metrics. This indicates that improving the effectiveness of supervision signals can accurately execute editing instructions while reducing disruption to other non-edited parts of the image, and preserving the quality and aesthetics of the original images. 
Specifically, we surpassed the current best methods by 3\%, 7\%, and 11\% Acc results in Following, Preserving, and Quality, respectively.

\vspace{-0.2cm}
\paragraph{Human Evaluation}
We also conduct a comprehensive human evaluation on Real-Edit benchmarks~\cite{multireward} in Tab.~\ref{tab:human_eval} and Fig.~\ref{fig:human_eval}. The assessment involved 15 experienced evaluators who rated edited images based on three critical metrics: instruction faithfulness (Following), preservation of irrelevant content (Preserving), and visual quality (Quality). The results of this manual evaluation demonstrate strong consistency with the GPT-4o scoring results shown in Tab.~\ref{tab:real_edit_benchmark}. This high alignment thoroughly validates that our proposed SuperEdit significantly outperforms existing methods across all evaluation criteria. Specifically, our SuperEdit surpasses the previous state-of-the-art method SmartEdit~\cite{smartedit} by 1.8\%, 16\%, 14.8\%, and 10.8\% on Following, Preserving, Quality, and Overall scores, respectively. These substantial improvements not only confirm the effectiveness of our approach but also establish SuperEdit as a new benchmark in instruction-guided image editing, achieving superior performance while requiring significantly less training data and cost.

\vspace{-0.2cm}
\paragraph{Visual Comparison with State-of-the-art Methods.}
We show the visual comparison with existing image editing methods in Fig.~\ref{fig:visual_comparison}. Compared to existing instruction-based editing methods, our approach not only better understands and executes editing instructions but also preserves the original image's layout and quality more effectively, thereby significantly outperforming previous methods. For example, with the instruction ``\textit{Replace the tiger with a lion, maintaining the same position in the water}" our SuperEdit method achieved superior results (4.8/4.8/4.8) compared to SmartEdit (4.8/4.8/2.5) and other methods.
Additionally, our method improves the model's comprehension of editing instructions. For the instruction ``\textit{Change the background to a sandy beach with the ocean in the distance}" our method received perfect scores (4.8/5.0/5.0) while SmartEdit only achieved (5.0/4.8/4.0). Similarly, for style transformation instructions like ``\textit{Change the image style to look like an impressionist painting style}" SuperEdit significantly outperformed SmartEdit with scores of (4.8/4.8/4.8) versus (1.0/4.8/4.8), demonstrating our method's superior ability to handle complex artistic transformations. Even more impressively, for scene transformation tasks like ``\textit{Transform the entire scene to a winter setting with snow covering the houses, trees, and boat}", our SuperEdit achieved (5.0/4.8/4.8) while SmartEdit only obtained (2.0/4.5/4.5). We provide more visual comparisons with other instruction-based image editing methods in the \textcolor{blue}{Supplementary Material.}

\subsection{Ablation Study}
\paragraph{Ablation on the Rectified and Contrastive Instructions.}
Considering that the Real-Edit~\cite{multireward} benchmark is evaluated by GPT-4o~\cite{gpt4}, and its evaluation results closely align with human ratings~\cite{multireward}, we choose this benchmark to conduct ablation experiments in Tab.~\ref{tab:ablation}. Compared to the original 300K InstructPix2Pix training data, our 40K training data with rectified editing instructions significantly improves all the performance of the editing model. Specifically, our approach improves scores by 0.95, 0.79, and 0.11, and accuracy by 21\%, 22\%, and 4\% in these three metrics, respectively.  In addition, editing performance is further enhanced by incorporating contrastive supervision signals. Compared to using only rectified editing instructions, the introduction of contrastive supervision signals improves the following and preserving scores by 0.19 and 0.08, and accuracy by 5\% and 2\%, while maintaining the quality accuracy and score. In summary, both the introduction of rectified editing instructions and contrastive editing instructions improve the overall performance of the editing model.

\begin{table}[t!]
\resizebox{1.0\linewidth}{!}{
\begin{tabular}{cc|ccccccc}
\specialrule{1.5pt}{0pt}{0pt}
\multirow{2}{*}{\begin{tabular}[c]{@{}c@{}}Rectified\\ Instruction\end{tabular}} & \multirow{2}{*}{\begin{tabular}[c]{@{}c@{}}Contrastive\\ Instructions\end{tabular}} & \multicolumn{2}{c}{Following$\uparrow$} & \multicolumn{2}{c}{Preserving$\uparrow$} & \multicolumn{2}{c}{Quality$\uparrow$} \\
&  & Acc & Score & Acc & Score & Acc & Score \\
 \hline
\xmark & \xmark & 41\% & 2.45 & 53\% & 3.27 & 61\% & 3.90 \\ 
\cmark & \xmark & 62\% & 3.40 & 75\% & 4.06 & 65\% & 4.01 \\
\cmark & \cmark & \textbf{67\%} & \textbf{3.59} & \textbf{77\%} & \textbf{4.14} & \textbf{65\%} & \textbf{4.01}
\\ \specialrule{1.5pt}{0pt}{0pt}
\end{tabular}
}
\vspace{-0.2cm}
\caption{Ablation study on our methods. Both rectified and contrastive editing instructions achieved improvements across all metrics.}
\vspace{-0.4cm}
\label{tab:ablation}
\end{table}

\vspace{-0.2cm}
\paragraph{Ablation on Data Scaling.} We investigated the impact of training data volume on model performance by experimenting with datasets ranging from 5k to 40k samples. Tab. \ref{tab:data_scaling} shows consistent improvements across all metrics as training data increases. With just 5k samples, our model achieves reasonable performance (54.7\% accuracy, 3.42 overall score), but scaling to 40k samples yields substantial gains (69.7\% accuracy, 3.91 overall score). The most significant improvements appear in the Preserving and Quality metrics, with 10\% and 15\%, respectively. This upward trend across all data points demonstrates that SuperEdit effectively leverages additional training examples without performance saturation, suggesting potential for further gains with larger datasets.
\begin{table}[th!]
\vspace{-0.15cm}
\resizebox{\linewidth}{!}{
\begin{tabular}{c|cccccccc}
\specialrule{1.5pt}{0pt}{0pt}
Data & \multicolumn{2}{c}{Following $\uparrow$} & \multicolumn{2}{c}{Preserving $\uparrow$} & \multicolumn{2}{c}{Quality $\uparrow$} & \multicolumn{2}{c}{Overall $\uparrow$} \\
 Size & Acc & Score & Acc & Score & Acc & Score & Acc & Score \\
\hline
5k & 49\% & 2.87 & 60\% & 3.71 & 55\% & 3.69 & 54.7\% & 3.42 \\
10k & 57\% & 3.26 & 71\% & 3.76 & 58\% & 3.87 & 62.0\% & 3.63 \\
20k & 64\% & 3.40 & 72\% & 4.02 & 63\% & 3.94 & 66.3\% & 3.79 \\
40k & \textbf{67\%} & \textbf{3.59} & \textbf{77\%} & \textbf{4.14} & \textbf{65\%} & \textbf{4.01} & \textbf{69.7\%} & \textbf{3.91} \\
\specialrule{1.5pt}{0pt}{0pt}
\end{tabular}
}
\vspace{-0.35cm}
\caption{Ablation study on data scaling results on Real-Edit~\cite{multireward}. }
\label{tab:data_scaling}
\vspace{-0.15cm}
\end{table}

\vspace{-0.2cm}
\section{Conclusion}
\label{sec:conclusion}
In this paper, we re-examine image editing models from the perspective of enhancing supervision signals, finding that existing models have not adequately addressed this challenge, resulting in suboptimal performance. We introduce a unified editing instruction rectification guideline based on diffusion priors that better aligns instructions with original-edited image pairs, thereby enhancing supervision effectiveness. We also construct contrastive editing instructions allowing models to learn from both positive and negative examples. Our data-oriented approach explores an important but overlooked research question: What level of performance can be achieved with minimal architectural modifications by primarily focusing on supervision quality and optimization? Remarkably, under both GPT-4o and human evaluation, our method outperforms existing approaches despite using less data and requiring no architectural modifications or additional pretraining. This shows high-quality supervision signals can effectively compensate for architectural simplicity, offering valuable new perspectives for image editing research.

\clearpage
{
    \small
    \bibliographystyle{ieeenat_fullname}
    \bibliography{main}

\begin{thebibliography}{62}
\providecommand{\natexlab}[1]{#1}
\providecommand{\url}[1]{\texttt{#1}}
\expandafter\ifx\csname urlstyle\endcsname\relax
  \providecommand{\doi}[1]{doi: #1}\else
  \providecommand{\doi}{doi: \begingroup \urlstyle{rm}\Url}\fi

\bibitem[Achiam et~al.(2023)Achiam, Adler, Agarwal, Ahmad, Akkaya, Aleman, Almeida, Altenschmidt, Altman, Anadkat, et~al.]{gpt4}
Josh Achiam, Steven Adler, Sandhini Agarwal, Lama Ahmad, Ilge Akkaya, Florencia~Leoni Aleman, Diogo Almeida, Janko Altenschmidt, Sam Altman, Shyamal Anadkat, et~al.
\newblock Gpt-4 technical report.
\newblock \emph{arXiv preprint arXiv:2303.08774}, 2023.

\bibitem[Balaji et~al.(2022)Balaji, Nah, Huang, Vahdat, Song, Zhang, Kreis, Aittala, Aila, Laine, et~al.]{ediff}
Yogesh Balaji, Seungjun Nah, Xun Huang, Arash Vahdat, Jiaming Song, Qinsheng Zhang, Karsten Kreis, Miika Aittala, Timo Aila, Samuli Laine, et~al.
\newblock ediff-i: Text-to-image diffusion models with an ensemble of expert denoisers.
\newblock \emph{arXiv preprint arXiv:2211.01324}, 2022.

\bibitem[Betker et~al.(2023)Betker, Goh, Jing, Brooks, Wang, Li, Ouyang, Zhuang, Lee, Guo, et~al.]{dalle3}
James Betker, Gabriel Goh, Li Jing, Tim Brooks, Jianfeng Wang, Linjie Li, Long Ouyang, Juntang Zhuang, Joyce Lee, Yufei Guo, et~al.
\newblock Improving image generation with better captions.
\newblock \emph{Computer Science. https://cdn. openai. com/papers/dall-e-3. pdf}, 2023.

\bibitem[Brooks et~al.(2023)Brooks, Holynski, and Efros]{instruct_p2p}
Tim Brooks, Aleksander Holynski, and Alexei~A Efros.
\newblock Instructpix2pix: Learning to follow image editing instructions.
\newblock In \emph{CVPR}, 2023.

\bibitem[Cao et~al.(2023)Cao, Wang, Qi, Shan, Qie, and Zheng]{masactrl}
Mingdeng Cao, Xintao Wang, Zhongang Qi, Ying Shan, Xiaohu Qie, and Yinqiang Zheng.
\newblock Masactrl: Tuning-free mutual self-attention control for consistent image synthesis and editing.
\newblock In \emph{ICCV}, 2023.

\bibitem[Couairon et~al.(2023)Couairon, Verbeek, Schwenk, and Cord]{diffedit}
Guillaume Couairon, Jakob Verbeek, Holger Schwenk, and Matthieu Cord.
\newblock Diffedit: Diffusion-based semantic image editing with mask guidance.
\newblock In \emph{ICLR}, 2023.

\bibitem[Dalva and Yanardag(2024)]{noise_clr}
Yusuf Dalva and Pinar Yanardag.
\newblock Noiseclr: A contrastive learning approach for unsupervised discovery of interpretable directions in diffusion models.
\newblock In \emph{CVPR}, 2024.

\bibitem[Deng et~al.(2024)Deng, Wang, Wei, Hou, and Grundmann]{prdp}
Fei Deng, Qifei Wang, Wei Wei, Tingbo Hou, and Matthias Grundmann.
\newblock Prdp: Proximal reward difference prediction for large-scale reward finetuning of diffusion models.
\newblock In \emph{CVPR}, 2024.

\bibitem[Dhariwal and Nichol(2021)]{diffusion_beat_gan}
Prafulla Dhariwal and Alexander Nichol.
\newblock Diffusion models beat gans on image synthesis.
\newblock \emph{NeurIPS}, 2021.

\bibitem[Esser et~al.(2024)Esser, Kulal, Blattmann, Entezari, M{\"u}ller, Saini, Levi, Lorenz, Sauer, Boesel, et~al.]{sd3}
Patrick Esser, Sumith Kulal, Andreas Blattmann, Rahim Entezari, Jonas M{\"u}ller, Harry Saini, Yam Levi, Dominik Lorenz, Axel Sauer, Frederic Boesel, et~al.
\newblock Scaling rectified flow transformers for high-resolution image synthesis.
\newblock In \emph{ICML}, 2024.

\bibitem[Fan et~al.(2024)Fan, Watkins, Du, Liu, Ryu, Boutilier, Abbeel, Ghavamzadeh, Lee, and Lee]{dpok}
Ying Fan, Olivia Watkins, Yuqing Du, Hao Liu, Moonkyung Ryu, Craig Boutilier, Pieter Abbeel, Mohammad Ghavamzadeh, Kangwook Lee, and Kimin Lee.
\newblock Reinforcement learning for fine-tuning text-to-image diffusion models.
\newblock \emph{NeurIPS}, 2024.

\bibitem[Fu et~al.(2024)Fu, Hu, Du, Wang, Yang, and Gan]{mgie}
Tsu-Jui Fu, Wenze Hu, Xianzhi Du, William~Yang Wang, Yinfei Yang, and Zhe Gan.
\newblock Guiding instruction-based image editing via multimodal large language models.
\newblock In \emph{ICLR}, 2024.

\bibitem[Ge et~al.(2024)Ge, Zhao, Li, Ge, and Shan]{seed_data_edit}
Yuying Ge, Sijie Zhao, Chen Li, Yixiao Ge, and Ying Shan.
\newblock Seed-data-edit technical report: A hybrid dataset for instructional image editing.
\newblock \emph{arXiv preprint arXiv:2405.04007}, 2024.

\bibitem[Geng et~al.(2024)Geng, Yang, Hang, Li, Gu, Zhang, Bao, Zhang, Li, Hu, et~al.]{instructdiffusion}
Zigang Geng, Binxin Yang, Tiankai Hang, Chen Li, Shuyang Gu, Ting Zhang, Jianmin Bao, Zheng Zhang, Houqiang Li, Han Hu, et~al.
\newblock Instructdiffusion: A generalist modeling interface for vision tasks.
\newblock In \emph{CVPR}, 2024.

\bibitem[Ghosh et~al.(2024)Ghosh, Hajishirzi, and Schmidt]{geneval}
Dhruba Ghosh, Hannaneh Hajishirzi, and Ludwig Schmidt.
\newblock Geneval: An object-focused framework for evaluating text-to-image alignment.
\newblock \emph{NeurIPS}, 2024.

\bibitem[Gu et~al.(2025)Gu, Li, Zhang, Chen, Wen, Luo, and Zhu]{multireward}
Xin Gu, Ming Li, Libo Zhang, Fan Chen, Longyin Wen, Tiejian Luo, and Sijie Zhu.
\newblock Multi-reward as condition for instruction-based image editing.
\newblock In \emph{ICLR}, 2025.

\bibitem[Guo and Lin(2024{\natexlab{a}})]{focus_instruction}
Qin Guo and Tianwei Lin.
\newblock Focus on your instruction: Fine-grained and multi-instruction image editing by attention modulation.
\newblock In \emph{CVPR}, 2024{\natexlab{a}}.

\bibitem[Guo and Lin(2024{\natexlab{b}})]{foi}
Qin Guo and Tianwei Lin.
\newblock Focus on your instruction: Fine-grained and multi-instruction image editing by attention modulation.
\newblock In \emph{CVPR}, 2024{\natexlab{b}}.

\bibitem[Hertz et~al.(2023)Hertz, Mokady, Tenenbaum, Aberman, Pritch, and Cohen-or]{p2p}
Amir Hertz, Ron Mokady, Jay Tenenbaum, Kfir Aberman, Yael Pritch, and Daniel Cohen-or.
\newblock Prompt-to-prompt image editing with cross-attention control.
\newblock In \emph{ICLR}, 2023.

\bibitem[Ho and Salimans(2022)]{cfg}
Jonathan Ho and Tim Salimans.
\newblock Classifier-free diffusion guidance.
\newblock \emph{arXiv preprint arXiv:2207.12598}, 2022.

\bibitem[Ho et~al.(2020)Ho, Jain, and Abbeel]{ddpm}
Jonathan Ho, Ajay Jain, and Pieter Abbeel.
\newblock Denoising diffusion probabilistic models.
\newblock \emph{NeurIPS}, 2020.

\bibitem[Hu et~al.(2024)Hu, Wang, Fang, Fu, Cheng, and Yu]{ella}
Xiwei Hu, Rui Wang, Yixiao Fang, Bin Fu, Pei Cheng, and Gang Yu.
\newblock Ella: Equip diffusion models with llm for enhanced semantic alignment.
\newblock \emph{arXiv preprint arXiv:2403.05135}, 2024.

\bibitem[Huang et~al.(2023)Huang, Sun, Xie, Li, and Liu]{T2i-compbench}
Kaiyi Huang, Kaiyue Sun, Enze Xie, Zhenguo Li, and Xihui Liu.
\newblock T2i-compbench: A comprehensive benchmark for open-world compositional text-to-image generation.
\newblock \emph{NeurIPS}, 2023.

\bibitem[Huang et~al.(2024)Huang, Xie, Wang, Yuan, Cun, Ge, Zhou, Dong, Huang, Zhang, et~al.]{smartedit}
Yuzhou Huang, Liangbin Xie, Xintao Wang, Ziyang Yuan, Xiaodong Cun, Yixiao Ge, Jiantao Zhou, Chao Dong, Rui Huang, Ruimao Zhang, et~al.
\newblock Smartedit: Exploring complex instruction-based image editing with multimodal large language models.
\newblock In \emph{CVPR}, 2024.

\bibitem[Hui et~al.(2024)Hui, Yang, Zhao, Shi, Wang, Wang, Zhou, and Xie]{hq_edit}
Mude Hui, Siwei Yang, Bingchen Zhao, Yichun Shi, Heng Wang, Peng Wang, Yuyin Zhou, and Cihang Xie.
\newblock Hq-edit: A high-quality dataset for instruction-based image editing.
\newblock \emph{arXiv preprint arXiv:2404.09990}, 2024.

\bibitem[Lee et~al.(2023)Lee, Phatale, Mansoor, Mesnard, Ferret, Lu, Bishop, Hall, Carbune, Rastogi, et~al.]{rlaif}
Harrison Lee, Samrat Phatale, Hassan Mansoor, Thomas Mesnard, Johan Ferret, Kellie Lu, Colton Bishop, Ethan Hall, Victor Carbune, Abhinav Rastogi, et~al.
\newblock Rlaif: Scaling reinforcement learning from human feedback with ai feedback.
\newblock \emph{arXiv preprint arXiv:2309.00267}, 2023.

\bibitem[Li et~al.(2024)Li, Yang, Kuang, Wu, Wang, Xiao, and Chen]{cpp}
Ming Li, Taojiannan Yang, Huafeng Kuang, Jie Wu, Zhaoning Wang, Xuefeng Xiao, and Chen Chen.
\newblock Controlnet++: Improving conditional controls with efficient consistency feedback.
\newblock In \emph{ECCV}, 2024.

\bibitem[Lin et~al.(2024)Lin, Chen, Tsai, Jiang, and Yang]{learnable_region}
Yuanze Lin, Yi-Wen Chen, Yi-Hsuan Tsai, Lu Jiang, and Ming-Hsuan Yang.
\newblock Text-driven image editing via learnable regions.
\newblock In \emph{CVPR}, 2024.

\bibitem[Liu et~al.(2024{\natexlab{a}})Liu, Wang, Cao, Jia, and Huang]{attn_edit}
Bingyan Liu, Chengyu Wang, Tingfeng Cao, Kui Jia, and Jun Huang.
\newblock Towards understanding cross and self-attention in stable diffusion for text-guided image editing.
\newblock In \emph{CVPR}, 2024{\natexlab{a}}.

\bibitem[Liu et~al.(2024{\natexlab{b}})Liu, Li, Wu, and Lee]{llava}
Haotian Liu, Chunyuan Li, Qingyang Wu, and Yong~Jae Lee.
\newblock Visual instruction tuning.
\newblock \emph{NeurIPS}, 2024{\natexlab{b}}.

\bibitem[Meng et~al.(2022)Meng, He, Song, Song, Wu, Zhu, and Ermon]{sdedit}
Chenlin Meng, Yutong He, Yang Song, Jiaming Song, Jiajun Wu, Jun-Yan Zhu, and Stefano Ermon.
\newblock Sdedit: Guided image synthesis and editing with stochastic differential equations.
\newblock In \emph{ICLR}, 2022.

\bibitem[Ouyang et~al.(2022)Ouyang, Wu, Jiang, Almeida, Wainwright, Mishkin, Zhang, Agarwal, Slama, Ray, et~al.]{instruct_gpt}
Long Ouyang, Jeffrey Wu, Xu Jiang, Diogo Almeida, Carroll Wainwright, Pamela Mishkin, Chong Zhang, Sandhini Agarwal, Katarina Slama, Alex Ray, et~al.
\newblock Training language models to follow instructions with human feedback.
\newblock \emph{NeurIPS}, 2022.

\bibitem[Pan et~al.(2024)Pan, Dong, Huang, Peng, Chen, and Wei]{Kosmos-G}
Xichen Pan, Li Dong, Shaohan Huang, Zhiliang Peng, Wenhu Chen, and Furu Wei.
\newblock Kosmos-g: Generating images in context with multimodal large language models.
\newblock In \emph{ICLR}, 2024.

\bibitem[Parmar et~al.(2023)Parmar, Kumar~Singh, Zhang, Li, Lu, and Zhu]{zero_i2i}
Gaurav Parmar, Krishna Kumar~Singh, Richard Zhang, Yijun Li, Jingwan Lu, and Jun-Yan Zhu.
\newblock Zero-shot image-to-image translation.
\newblock In \emph{ACM SIGGRAPH}, 2023.

\bibitem[Paszke et~al.(2019)Paszke, Gross, Massa, Lerer, Bradbury, Chanan, Killeen, Lin, Gimelshein, Antiga, et~al.]{pytorch}
Adam Paszke, Sam Gross, Francisco Massa, Adam Lerer, James Bradbury, Gregory Chanan, Trevor Killeen, Zeming Lin, Natalia Gimelshein, Luca Antiga, et~al.
\newblock Pytorch: An imperative style, high-performance deep learning library.
\newblock \emph{NeurIPS}, 2019.

\bibitem[Patashnik et~al.(2023)Patashnik, Garibi, Azuri, Averbuch-Elor, and Cohen-Or]{localizing_diffusion}
Or Patashnik, Daniel Garibi, Idan Azuri, Hadar Averbuch-Elor, and Daniel Cohen-Or.
\newblock Localizing object-level shape variations with text-to-image diffusion models.
\newblock In \emph{ICCV}, 2023.

\bibitem[Podell et~al.(2023)Podell, English, Lacey, Blattmann, Dockhorn, M{\"u}ller, Penna, and Rombach]{sdxl}
Dustin Podell, Zion English, Kyle Lacey, Andreas Blattmann, Tim Dockhorn, Jonas M{\"u}ller, Joe Penna, and Robin Rombach.
\newblock Sdxl: Improving latent diffusion models for high-resolution image synthesis.
\newblock In \emph{ICLR}, 2023.

\bibitem[Radford et~al.(2021)Radford, Kim, Hallacy, Ramesh, Goh, Agarwal, Sastry, Askell, Mishkin, Clark, et~al.]{clip}
Alec Radford, Jong~Wook Kim, Chris Hallacy, Aditya Ramesh, Gabriel Goh, Sandhini Agarwal, Girish Sastry, Amanda Askell, Pamela Mishkin, Jack Clark, et~al.
\newblock Learning transferable visual models from natural language supervision.
\newblock In \emph{ICML}, 2021.

\bibitem[Rafailov et~al.(2024)Rafailov, Sharma, Mitchell, Manning, Ermon, and Finn]{dpo}
Rafael Rafailov, Archit Sharma, Eric Mitchell, Christopher~D Manning, Stefano Ermon, and Chelsea Finn.
\newblock Direct preference optimization: Your language model is secretly a reward model.
\newblock \emph{NeurIPS}, 2024.

\bibitem[Ramesh et~al.(2021)Ramesh, Pavlov, Goh, Gray, Voss, Radford, Chen, and Sutskever]{dalle}
Aditya Ramesh, Mikhail Pavlov, Gabriel Goh, Scott Gray, Chelsea Voss, Alec Radford, Mark Chen, and Ilya Sutskever.
\newblock Zero-shot text-to-image generation.
\newblock In \emph{ICML}, 2021.

\bibitem[Ramesh et~al.(2022)Ramesh, Dhariwal, Nichol, Chu, and Chen]{dalle2}
Aditya Ramesh, Prafulla Dhariwal, Alex Nichol, Casey Chu, and Mark Chen.
\newblock Hierarchical text-conditional image generation with clip latents.
\newblock \emph{arXiv preprint arXiv:2204.06125}, 2022.

\bibitem[Rombach et~al.(2022)Rombach, Blattmann, Lorenz, Esser, and Ommer]{sd}
Robin Rombach, Andreas Blattmann, Dominik Lorenz, Patrick Esser, and Bj{\"o}rn Ommer.
\newblock High-resolution image synthesis with latent diffusion models.
\newblock In \emph{CVPR}, 2022.

\bibitem[Ronneberger et~al.(2015)Ronneberger, Fischer, and Brox]{unet}
Olaf Ronneberger, Philipp Fischer, and Thomas Brox.
\newblock U-net: Convolutional networks for biomedical image segmentation.
\newblock In \emph{MICCAI}, 2015.

\bibitem[Saharia et~al.(2022)Saharia, Chan, Saxena, Li, Whang, Denton, Ghasemipour, Gontijo~Lopes, Karagol~Ayan, Salimans, et~al.]{imagen}
Chitwan Saharia, William Chan, Saurabh Saxena, Lala Li, Jay Whang, Emily~L Denton, Kamyar Ghasemipour, Raphael Gontijo~Lopes, Burcu Karagol~Ayan, Tim Salimans, et~al.
\newblock Photorealistic text-to-image diffusion models with deep language understanding.
\newblock \emph{NeurIPS}, 2022.

\bibitem[Sheynin et~al.(2024)Sheynin, Polyak, Singer, Kirstain, Zohar, Ashual, Parikh, and Taigman]{emu_edit}
Shelly Sheynin, Adam Polyak, Uriel Singer, Yuval Kirstain, Amit Zohar, Oron Ashual, Devi Parikh, and Yaniv Taigman.
\newblock Emu edit: Precise image editing via recognition and generation tasks.
\newblock In \emph{CVPR}, 2024.

\bibitem[Simsar et~al.(2023)Simsar, Tonioni, Xian, Hofmann, and Tombari]{lime}
Enis Simsar, Alessio Tonioni, Yongqin Xian, Thomas Hofmann, and Federico Tombari.
\newblock Lime: localized image editing via attention regularization in diffusion models.
\newblock \emph{arXiv preprint arXiv:2312.09256}, 2023.

\bibitem[Singh et~al.(2024)Singh, Zhang, Liu, Smith, Lin, and Zheng]{smartmask}
Jaskirat Singh, Jianming Zhang, Qing Liu, Cameron Smith, Zhe Lin, and Liang Zheng.
\newblock Smartmask: Context aware high-fidelity mask generation for fine-grained object insertion and layout control.
\newblock In \emph{CVPR}, 2024.

\bibitem[Sohl-Dickstein et~al.(2015)Sohl-Dickstein, Weiss, Maheswaranathan, and Ganguli]{diffusion}
Jascha Sohl-Dickstein, Eric Weiss, Niru Maheswaranathan, and Surya Ganguli.
\newblock Deep unsupervised learning using nonequilibrium thermodynamics.
\newblock In \emph{ICML}, 2015.

\bibitem[Song et~al.(2021)Song, Meng, and Ermon]{ddim}
Jiaming Song, Chenlin Meng, and Stefano Ermon.
\newblock Denoising diffusion implicit models.
\newblock In \emph{ICLR}, 2021.

\bibitem[Team(2024)]{kolors}
Kolors Team.
\newblock Kolors: Effective training of diffusion model for photorealistic text-to-image synthesis.
\newblock \emph{arXiv preprint}, 2024.

\bibitem[Tumanyan et~al.(2023)Tumanyan, Geyer, Bagon, and Dekel]{pnp}
Narek Tumanyan, Michal Geyer, Shai Bagon, and Tali Dekel.
\newblock Plug-and-play diffusion features for text-driven image-to-image translation.
\newblock In \emph{CVPR}, 2023.

\bibitem[Wallace et~al.(2024)Wallace, Dang, Rafailov, Zhou, Lou, Purushwalkam, Ermon, Xiong, Joty, and Naik]{diffusion_dpo}
Bram Wallace, Meihua Dang, Rafael Rafailov, Linqi Zhou, Aaron Lou, Senthil Purushwalkam, Stefano Ermon, Caiming Xiong, Shafiq Joty, and Nikhil Naik.
\newblock Diffusion model alignment using direct preference optimization.
\newblock In \emph{CVPR}, 2024.

\bibitem[Xie et~al.(2023)Xie, Zhang, Lin, Hinz, and Zhang]{smartbrush}
Shaoan Xie, Zhifei Zhang, Zhe Lin, Tobias Hinz, and Kun Zhang.
\newblock Smartbrush: Text and shape guided object inpainting with diffusion model.
\newblock In \emph{CVPR}, 2023.

\bibitem[Xu et~al.(2023)Xu, Liu, Wu, Tong, Li, Ding, Tang, and Dong]{imagereward}
Jiazheng Xu, Xiao Liu, Yuchen Wu, Yuxuan Tong, Qinkai Li, Ming Ding, Jie Tang, and Yuxiao Dong.
\newblock Imagereward: Learning and evaluating human preferences for text-to-image generation.
\newblock \emph{NeurIPS}, 2023.

\bibitem[Yang et~al.(2024)Yang, Tao, Lyu, Ge, Chen, Shen, Zhu, and Li]{D3PO}
Kai Yang, Jian Tao, Jiafei Lyu, Chunjiang Ge, Jiaxin Chen, Weihan Shen, Xiaolong Zhu, and Xiu Li.
\newblock Using human feedback to fine-tune diffusion models without any reward model.
\newblock In \emph{CVPR}, 2024.

\bibitem[Yi et~al.(2024)Yi, Li, Xin, and Li]{understanding_diffusion}
Mingyang Yi, Aoxue Li, Yi Xin, and Zhenguo Li.
\newblock Towards understanding the working mechanism of text-to-image diffusion model.
\newblock \emph{arXiv preprint arXiv:2405.15330}, 2024.

\bibitem[Yu et~al.(2023)Yu, Feng, Feng, Liu, Jin, Zeng, and Chen]{inpaint_anything}
Tao Yu, Runseng Feng, Ruoyu Feng, Jinming Liu, Xin Jin, Wenjun Zeng, and Zhibo Chen.
\newblock Inpaint anything: Segment anything meets image inpainting.
\newblock \emph{arXiv preprint arXiv:2304.06790}, 2023.

\bibitem[Zhang et~al.(2024{\natexlab{a}})Zhang, Mo, Chen, Sun, and Su]{magicbrush}
Kai Zhang, Lingbo Mo, Wenhu Chen, Huan Sun, and Yu Su.
\newblock Magicbrush: A manually annotated dataset for instruction-guided image editing.
\newblock \emph{NeurIPS}, 2024{\natexlab{a}}.

\bibitem[Zhang et~al.(2023)Zhang, Rao, and Agrawala]{controlnet}
Lvmin Zhang, Anyi Rao, and Maneesh Agrawala.
\newblock Adding conditional control to text-to-image diffusion models.
\newblock In \emph{ICCV}, 2023.

\bibitem[Zhang et~al.(2024{\natexlab{b}})Zhang, Yang, Feng, Qin, Chen, Yu, Chen, Wang, Savarese, Ermon, et~al.]{hive}
Shu Zhang, Xinyi Yang, Yihao Feng, Can Qin, Chia-Chih Chen, Ning Yu, Zeyuan Chen, Huan Wang, Silvio Savarese, Stefano Ermon, et~al.
\newblock Hive: Harnessing human feedback for instructional visual editing.
\newblock In \emph{CVPR}, 2024{\natexlab{b}}.

\bibitem[Zhang et~al.(2024{\natexlab{c}})Zhang, Liu, Xie, Faccio, Shou, and Schmidhuber]{tgate}
Wentian Zhang, Haozhe Liu, Jinheng Xie, Francesco Faccio, Mike~Zheng Shou, and J{\"u}rgen Schmidhuber.
\newblock Cross-attention makes inference cumbersome in text-to-image diffusion models.
\newblock \emph{arXiv preprint arXiv:2404.02747}, 2024{\natexlab{c}}.

\bibitem[Zhao et~al.(2024)Zhao, Ma, Chen, Si, Wu, An, Yu, Zhang, Li, and Chang]{ultraedit}
Haozhe Zhao, Xiaojian Ma, Liang Chen, Shuzheng Si, Rujie Wu, Kaikai An, Peiyu Yu, Minjia Zhang, Qing Li, and Baobao Chang.
\newblock Ultraedit: Instruction-based fine-grained image editing at scale.
\newblock \emph{arXiv preprint arXiv:2407.05282}, 2024.

\end{thebibliography}
}

\clearpage
\setcounter{page}{1}
\maketitlesupplementary

\section{Overview of Supplementary}
The supplementary material is organized into the following sections:
\begin{itemize}
    \item Section~\ref{implementation}: Implementation details.
    \item Section~\ref{supp_experiments}: More experiments and analysis.
    \item Section~\ref{supp_diffusion_prior}: More analysis on diffusion generation prior.
    \item Section~\ref{supp_prompt}: Detailed prompt for generation prior.
    \item Section~\ref{supp_discussion}: Discussion and limitation.
    \item Section~\ref{supp_visualization}: More visualization comparison and results.
\end{itemize}

\section{Implementation Details}
\label{implementation}
We implemented our editing model training based on the InstructPix2Pix PyTorch~\cite{pytorch} code from the Diffusers repository~\cite{diffusion}, using Stable Diffusion V1.5~\cite{sd} as the pre-trained weights for the editing model. Following InstructPix2Pix's implementation~\cite{instruct_p2p}, we enable classiﬁer-free diffusion guidance~\cite{cfg} for both the image condition and the text condition with 5\% mask probability during training. The training batch size is 512 with a learning rate of 1e-4, weight decay of 1e-2, and a warm-up ratio of 100 steps. The training resolution is 512x512 by resizing input images without any crops. Margin $m=5e-3$ and weight $\lambda=1.0$ is used for triplet loss $\mathcal{L}_{\text {triplet}}$. We train the edit model for 10,000 steps and use the triplet loss after the 2,000 training steps. During inference, we keep the original image ratio and resize the shorter side to 512, with DDIM~\cite{ddim} sampler and 50 sampling steps, following the default settings of Multi-Reward~\cite{multireward}. The text guidance scale and image guidance scale we used for inference are 10.0 and 1.5, respectively.

\section{More Experiments and Analysis}
In this section, we provide more experiments and analysis. We first discuss limitations of current metrics in Sec.~\ref{metric_limitation}, then present the MagicBrush benchmark results in Sec.~\ref{magicbrush_benchmark}, and finally analyze GPT-4o cost and different VLMs in Sec.~\ref{gpt_cost}.

\label{supp_experiments}
\subsection{Limitations of Existing Metrics}
\label{metric_limitation}
\begin{figure}[h]\centering
    \vspace{-0.2cm}
    \includegraphics[width=1.0\linewidth]{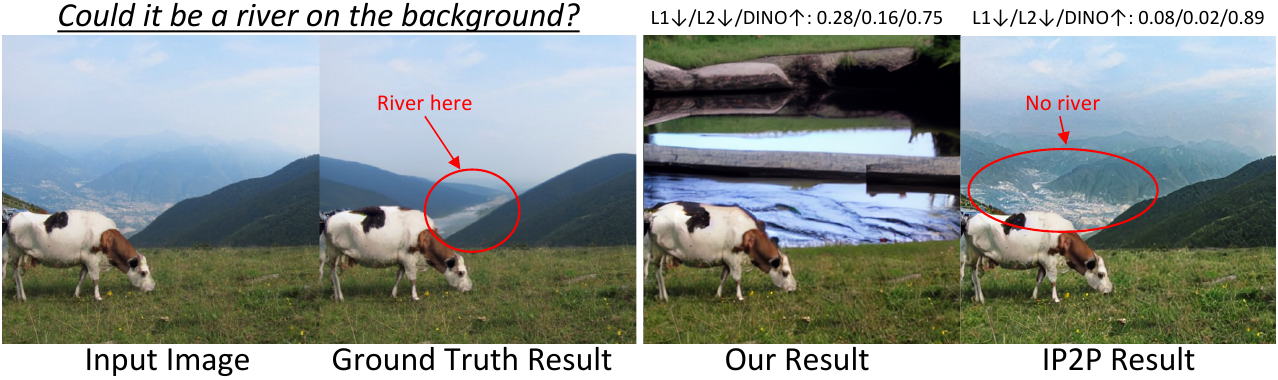}
    \vspace{-0.7cm}
    \caption{Existing metrics cannot reliably indicate editing quality.}
    \vspace{-0.4cm}
    \label{fig:metric_issue}
\end{figure}

Here, we show an example from MagicBrush test set in Fig.~\ref{fig:metric_issue} to illustrate that existing metrics (e.g., L1/L2/DINO) cannot reflect actual editing quality; that is, the results of these metrics do not match human judgment. This dilemma has also been noted in previous instruction-based image editing works, including SmartEdit~\cite{smartedit}, Emu-Edit~\cite{emu_edit}, and MultiReward~\cite{multireward}. 

In addition, SmartEdit's metrics (CLIP, DINO) in Tab.~\ref{tab:mb_benchmark} are worse than MagicBrush, but its human evaluation shows better results in SmartEdit paper~\cite{smartedit}. This discrepancy further shows the rationale for our comprehensive assessment using both GPT-4o-based evaluation (RealEdit) and human evaluation. 

\subsection{Evaluation on  MagicBrush Benchmark}
\label{magicbrush_benchmark}
In Tab.~\ref{tab:mb_benchmark}, we present a quantitative comparison of various image editing methods evaluated on the MagicBrush single-turn benchmark. However, it's important to note that these automated metrics (CLIP-I, CLIP-T, DINO, L1) should be interpreted with caution. As highlighted by previous works~\cite{smartedit,emu_edit,multireward}, such metrics often fail to fully capture human perceptual preferences, and can sometimes lead to misleading conclusions about actual editing quality. Several studies have demonstrated significant discrepancies between metric-based rankings and human evaluation results~\cite{smartedit,emu_edit,multireward}.

Our proposed method adopts a data-oriented approach, contrasting with the model-oriented strategies prevalent in image editing. Remarkably, without requiring additional parameters, pretraining tasks, or extensive training data (using only 40K samples compared to 300K-1.2M in other methods), our approach achieves competitive performance across all metrics. The CLIP-T score of 30.3 is only 0.3 lower than the best results, and DINO score of 80.2 (second highest) is particularly noteworthy, suggesting strong preservation of both semantic and structural image features.

\begin{table}[h]
\vspace{-0.2cm}
\centering
\resizebox{\linewidth}{!}{
  \begin{tabular}{c|ccc|cccc}
  \specialrule{1.5pt}{0pt}{0pt}
  Method & \begin{tabular}[c]{@{}c@{}}Extra\\ Module\end{tabular} & \begin{tabular}[c]{@{}c@{}}Pretrain\\ Tasks\end{tabular} & \begin{tabular}[c]{@{}c@{}}Edit\\ Data\end{tabular} & CLIP-I$\uparrow$ & CLIP-T$\uparrow$ & DINO$\uparrow$ & L1$\downarrow$ \\
  \hline
  InstructPix2Pix~\cite{instruct_p2p} & \xmark & \xmark & 300K & 85.4 & 29.2 & 69.8 & 0.112 \\
  InstructDiffusion~\cite{instructdiffusion} & \xmark & \cmark & 860K & 89.2 & 30.2 & 77.7 & -\\
  MagicBrush~\cite{magicbrush} & \xmark & \xmark & 310K & \textbf{90.7} & \textbf{30.6} & \textbf{80.6} & \textbf{0.062} \\
  SmartEdit~\cite{smartedit} & \cmark & \cmark & 1.2M & 90.4 & 30.3 & 79.7 & \uline{0.081} \\
  SuperEdit (Ours) & \xmark & \xmark & 40K & \uline{90.5}  & \uline{30.3} & \uline{80.2} & 0.106
  \\
  \specialrule{1.5pt}{0pt}{0pt}
  \end{tabular}
  \vspace{-0.6cm}
}
\vspace{-0.3cm}
\caption{Quantitative comparison (L1/CLIP-I/CLIP-T/DINO-I) on the MagicBrush benchmark. Our SuperEdit achieves good performance with better efficiency, without extra modules or pretrain tasks.}
\vspace{-0.4cm}
\label{tab:mb_benchmark}
\end{table}

\begin{figure*}[th!]\centering
    \includegraphics[width=1.0\linewidth]{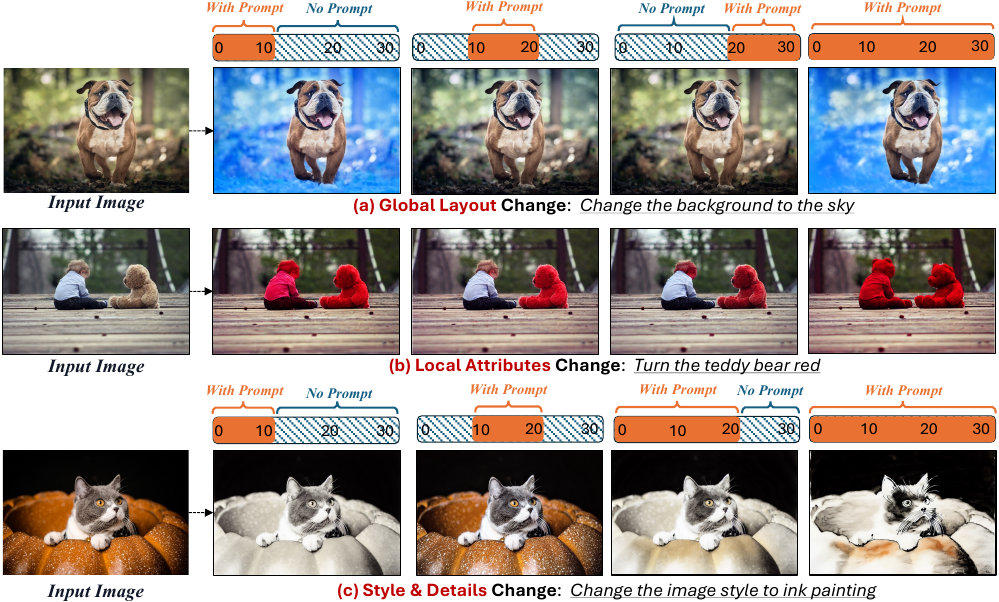}
    \vspace{-0.6cm}
    \caption{
    We show the impact of incorporating the editing prompt at different inference timesteps on the edited image. \textbf{(a)} The global layout changes usually occur in the early stages of inference. Adding text editing instructions to modify the global layout at the mid or late stages does not effectively impact the global layout. \textbf{(b)} Local object attribute changes occur in the mid-stages of sampling. Adding text editing instructions in the early or late stages may result in incorrect editing outcomes. \textbf{(c)} The style changes happen across all inference stages, and the detail changes happen in the late stage (Please refer to the subtle differences between the last two images). Best viewed in color.
    }
    \label{fig:supp_diffusion_prior}
\end{figure*}

\subsection{GPT-4o Cost \& Different VLMs' Performance}
\label{gpt_cost}
We respectfully emphasize that our core contribution is identifying and addressing noisy supervision in existing datasets, rather than focusing on cost-effective scaling strategies. Using GPT-4o for our method costs \$0.02 per 512×512 input-edited image pair, totaling \$800 for 40K data, which is less expensive than existing works that require additional VLM fine-tuning or extra pre-training stages. For alternative ablation, we asked 5 annotators to evaluate rectified instructions from different VLMs. As shown in Tab.~\ref{tab:alternatives}, existing open-source VLMs can partially substitute GPT-4o. These open-source models can be further fine-tuned with GPT-4o data and then used for efficient scaling up, which we leave for future work.

\vspace{-0.1cm}
\begin{table}[h]
\resizebox{\columnwidth}{!}{
\begin{tabular}{cccc}
\hline
GPT-4o & LLaVA-OV(72B) & InternVL2(76B) & Qwen2-VL(72B)
\\ \hline
\textbf{76.2\%} & 50.4\% & 48.2\% & 47.8\% 
\\ \hline
\end{tabular}
}
\vspace{-0.4cm}
\caption{Instruction rectification success rate across 100 samples}
\vspace{-0.4cm}
\label{tab:alternatives}
\end{table}

\section{Diffusion Generation Prior}
\vspace{-0.2cm}
\label{supp_diffusion_prior}
As discussed in Sec.~\ref{diffusion_prior} and Fig.~\ref{fig:diffusion_prior} of the main paper, editing diffusion models focus on specific generation attributes during inference, independent of the different editing instructions. Specifically, editing models focus on global layout in the early stages, local object attributes in the mid stages, image details in the late stages, and style change across all sampling stages. In this section, we further demonstrate this generation prior in Fig.\ref{fig:supp_diffusion_prior}.

Fig.\ref{fig:supp_diffusion_prior} provides compelling visual evidence for the claims made in the main paper regarding how diffusion models process different aspects of image generation at specific timesteps. The experiments systematically demonstrate that this behavior is consistent across various editing tasks, reinforcing the observation that ``different timesteps play distinct roles in image generation for text-to-image diffusion models, regardless of the text prompt" as cited in previous works.

Specifically, the figure illustrates three key patterns:
(a) Global Layout Changes: The first row shows that changing the background to sky is most effective when prompts are introduced in the early stages (0-10 timesteps). When the same editing instruction is applied during mid (10-20) or late (20-30) stages, the model fails to properly modify the global layout, maintaining the original forest background despite the editing instructions. This validates our assertion that ``diffusion models focus on global layout in the early stages."
(b) Local Object Attributes: The second row demonstrates that local attribute modifications, such as changing the teddy bear's color to red, are optimally achieved during the mid-stages of sampling (10-20 timesteps). When the color change instruction is introduced too early or too late, the results show inconsistent or incomplete color transformation. This confirms that ``local object attributes are processed in the mid stages".
(c) Style and Details: The third row reveals two important insights. First, style transformations (changing to ink painting style) can be effectively applied across all timesteps, indicating that style modifications have a more flexible temporal window. Second, subtle detail refinements are predominantly processed in the late stages (20-30), as evidenced by the finer differences between the last two images in the bottom row. This supports our claim about ``image details in the late stages of sampling."
These observations not only validate the theoretical framework presented in the main text but also provide practical insights for optimizing instruction-based image editing. The clear temporal division of editing capabilities suggests that a more nuanced approach to prompt timing could significantly improve editing outcomes. This understanding directly supports our approach of guiding Vision-Language Models based on these four generation attributes (global layout, local attributes, style, and details), enabling us to establish a unified rectification method applicable across various editing instructions as described in the main paper.

\section{GPT-4o Prompts for Constructing Rectified and Contrastive Editing Instructions}
\label{supp_prompt}
\begin{figure*}[t!]\centering
    \includegraphics[width=0.9\linewidth]{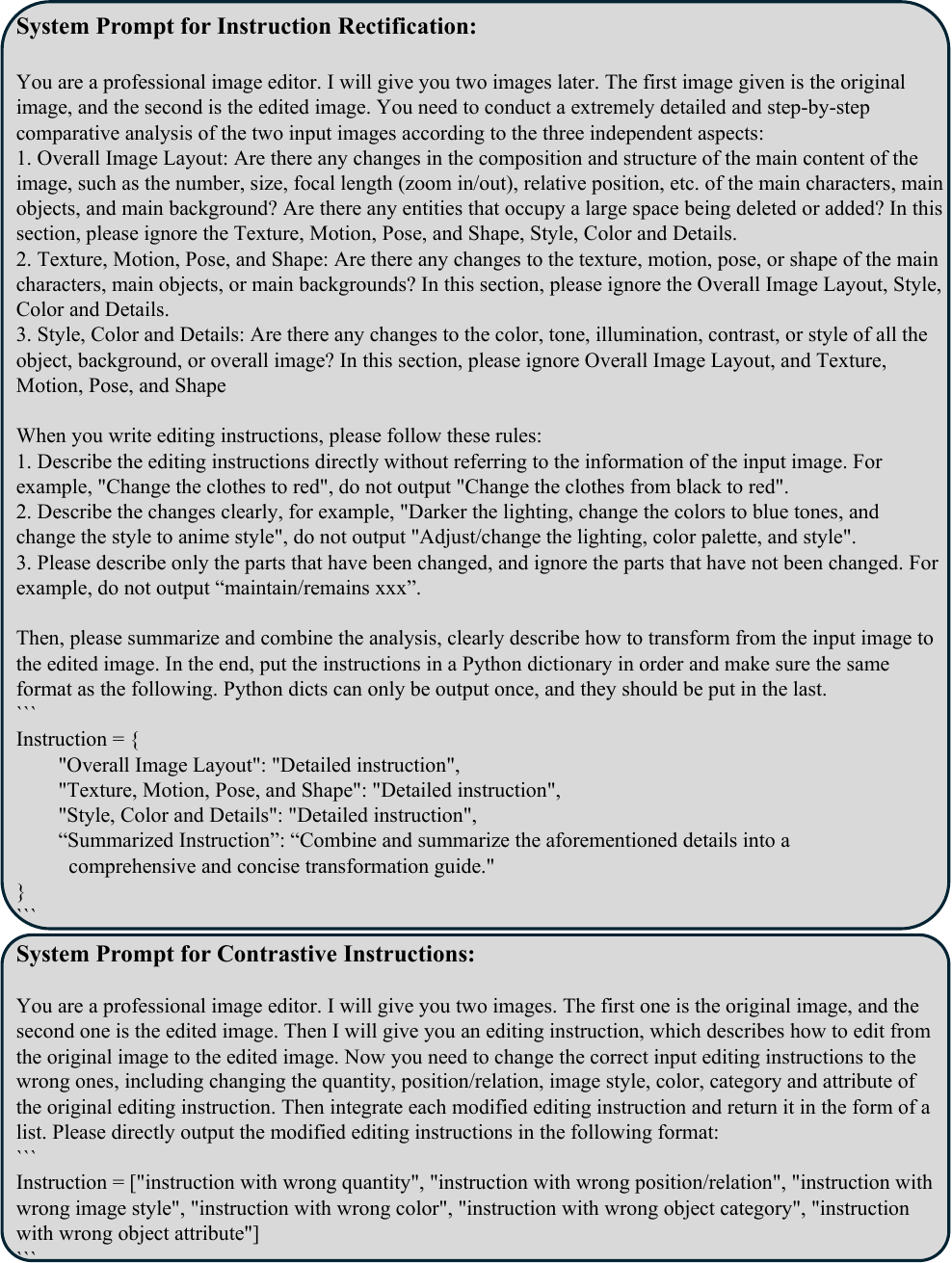}
    \caption{
    GPT-4o prompts for constructing rectified and contrastive editing instructions
    }
    \label{fig:gpt_prompts}
\end{figure*}

\vspace{-0.2cm}
We show the detailed prompt for GPT-4o to construct the rectified and contrastive editing instructions in Fig.~\ref{fig:gpt_prompts}.
As discussed in Sec.~\ref{supp_diffusion_prior}, we input the original image and the edited image into GPT-4o and ask it to return the differences in the following four attributes: ``Overall Image Layout ``Local Object Attributes", ``Image Details", and ``Style Change". When calling the GPT-4o API, we explicitly define ``Overall Image Layout" as modifications to the major objects, characters, and background in the image. ``Local Object Attributes" are defined as changes in the texture, motion, pose, and shape of the major objects, characters, and background. Additionally, we combine ``Style" and ``Details" into a single category to reduce the number of tokens generated by GPT-4o, thus saving costs. We observed that this adjustment does not reduce GPT-4o's understanding of the style and detail changes between the original-edited image pair. In the actual training of the editing model, acknowledging that CLIP~\cite{clip} text encoder can accept a maximum of 77 textual tokens as input, we ask GPT-4o to summarize and refine these rectified instructions. We then use the consolidated and refined editing instructions (``Summarized Instruction" in Fig.~\ref{fig:gpt_prompts}) to train the model.

\section{Discussion and Limitation}
\label{supp_discussion}
\vspace{-0.2cm}

\paragraph{Discussion.}
It's important to emphasize that our data-oriented approach is not mutually exclusive with model-oriented methods like MultiReward or SmartEdit, nor is its purpose to surpass existing work across various benchmarks or diminish their excellent contributions. Instead, our work explores a complementary yet important research question: What level of performance can be achieved with minimal architectural modifications by primarily focusing on supervision quality and optimization? Surprisingly, under both GPT-4o and human evaluation, our method significantly outperforms existing approaches despite using only a small amount of data, without modifying the model architecture, and requiring no additional pretraining. This suggests that high-quality data can substantially compensate for architectural simplicity, achieving results comparable to or even better than methods with considerably more parameters and pretraining requirements. We believe our approach and experimental results bring new insights and novelty to the field of image editing research.

Furthermore, since our data-oriented approach is complementary and orthogonal to existing work, we can build upon current methods to further improve editing performance. Specifically, we follow the same setup as SmartEdit, retraining our model using InstructDiffusion as the pre-trained weights. The experimental results, as shown in Tab.~\ref{tab:supp_instructdiffusion_pretrain}, demonstrate that our method can complement existing work to achieve even better editing performance. When comparing SuperEdit with InstructDiffusion pre-trained weights against SmartEdit, we observe significant improvements across all metrics (71\% vs. 64\% in following instructions, 83\% vs. 66\% in preserving content, and 71\% vs. 45\% in image quality), despite using only 40K training samples compared to SmartEdit's 1.2M.

\begin{table}[h]
    \centering
    \vspace{-0.3cm}
    \resizebox{\columnwidth}{!}{
    \begin{tabular}{c|cc|cccccc}
    \specialrule{1.5pt}{0pt}{0pt}
    \multirow{2}{*}{Method} & \multirow{2}{*}{\begin{tabular}[c]{@{}c@{}}Pre-trained\\ U-Net\end{tabular}} & \multirow{2}{*}{\begin{tabular}[c]{@{}c@{}}Model Size\\ Edit Data\end{tabular}} & \multicolumn{2}{c}{Following $\uparrow$} & \multicolumn{2}{c}{Preserving $\uparrow$} & \multicolumn{2}{c}{Quality $\uparrow$} \\
     & & & Acc & Score & Acc & Score & Acc & Score \\
     \hline
     SmartEdit & InstrutDiff & 14.1B/1.2M & 64\% & 3.50 & 66\% & 3.70 & 45\% & 3.56 \\
     \hline
     SuperEdit & SD1.5 & 1.1B/40K & 67\% & 3.59 & 77\% & 4.14 & 65\% & 4.01 \\
     SuperEdit & InstrutDiff & 1.1B/40K & \textbf{71\%} & \textbf{3.76} & \textbf{83\%} & \textbf{4.32} & \textbf{71\%} & \textbf{4.17}
    \\ \specialrule{1.5pt}{0pt}{0pt}
    \end{tabular}
    }
    \vspace{-0.3cm}
    \caption{SuperEdit outperforms previous SOTA SmartEdit and achieves further improvements with InstructDiffusion pre-trained weights.}
    \vspace{-0.4cm}
    \label{tab:supp_instructdiffusion_pretrain}
\end{table}

In addition, we also provide the results that trained with a lower resolution (256 $\times$ 256), the results on Real-Edit benchmark still outperforms previous SOTA method SmartEdit~\cite{smartedit}.

\begin{table}[h]
    \centering
    \vspace{-0.3cm}
    \resizebox{\columnwidth}{!}{
    \begin{tabular}{c|cc|cccccc}
    \specialrule{1.5pt}{0pt}{0pt}
    \multirow{2}{*}{Method} & \multirow{2}{*}{\begin{tabular}[c]{@{}c@{}}Model Size\\ Edit Data\end{tabular}} & \multirow{2}{*}{\begin{tabular}[c]{@{}c@{}}Training\\ Resolution\end{tabular}} & \multicolumn{2}{c}{Following $\uparrow$} & \multicolumn{2}{c}{Preserving $\uparrow$} & \multicolumn{2}{c}{Quality $\uparrow$} \\
     & & & Acc & Score & Acc & Score & Acc & Score \\
     \hline
     SmartEdit & 14.1B/1.2M & 256 & 64\% & 3.50 & 66\% & 3.70 & 45\% & 3.56 \\
     \hline
     SuperEdit & 1.1B/40K & 256 & \textbf{68\%} & \textbf{3.56} & \textbf{75\%} & \textbf{4.02} & \textbf{66\%} & \textbf{4.02} \\
    \specialrule{1.5pt}{0pt}{0pt}
    \end{tabular}
    }
    \vspace{-0.2cm}
    \caption{SuperEdit results with lower training resolution. Both SmartEdit and SuperEdit are pre-trained with InstructDiffusion here.}
    \vspace{-0.8cm}
    \label{tab:supp_diff_resolution}
\end{table}

\paragraph{Limitation.}
Our method significantly enhances instruction-based image editing, but limitations still exist. The trained model still faces difficulties in understanding and executing complex instructions, especially with densely arranged objects and complicated spatial relationships. Although we used correction instructions and contrastive supervision signals, differences between editing results and editing instructions may still occur due to the inherent limitations of pre-trained Stable Diffusion and the challenges in fully capturing the nuances of natural language. Additionally, to fairly compare with existing methods, we chose Stable Diffusion v1.5 as the Base Model for building our editing model, which may result in worse image quality of edited images compared to state-of-the-art Text-to-Image models. Finally, ensuring the accuracy and effectiveness of correction and contrastive instructions requires the use of GPT-4o~\cite{gpt4}, which may incur additional costs as the amount of data increases.

\section{More Visualization Comparison and Results}
\label{supp_visualization}
\begin{figure*}[t!]\centering
    \includegraphics[width=\textwidth]{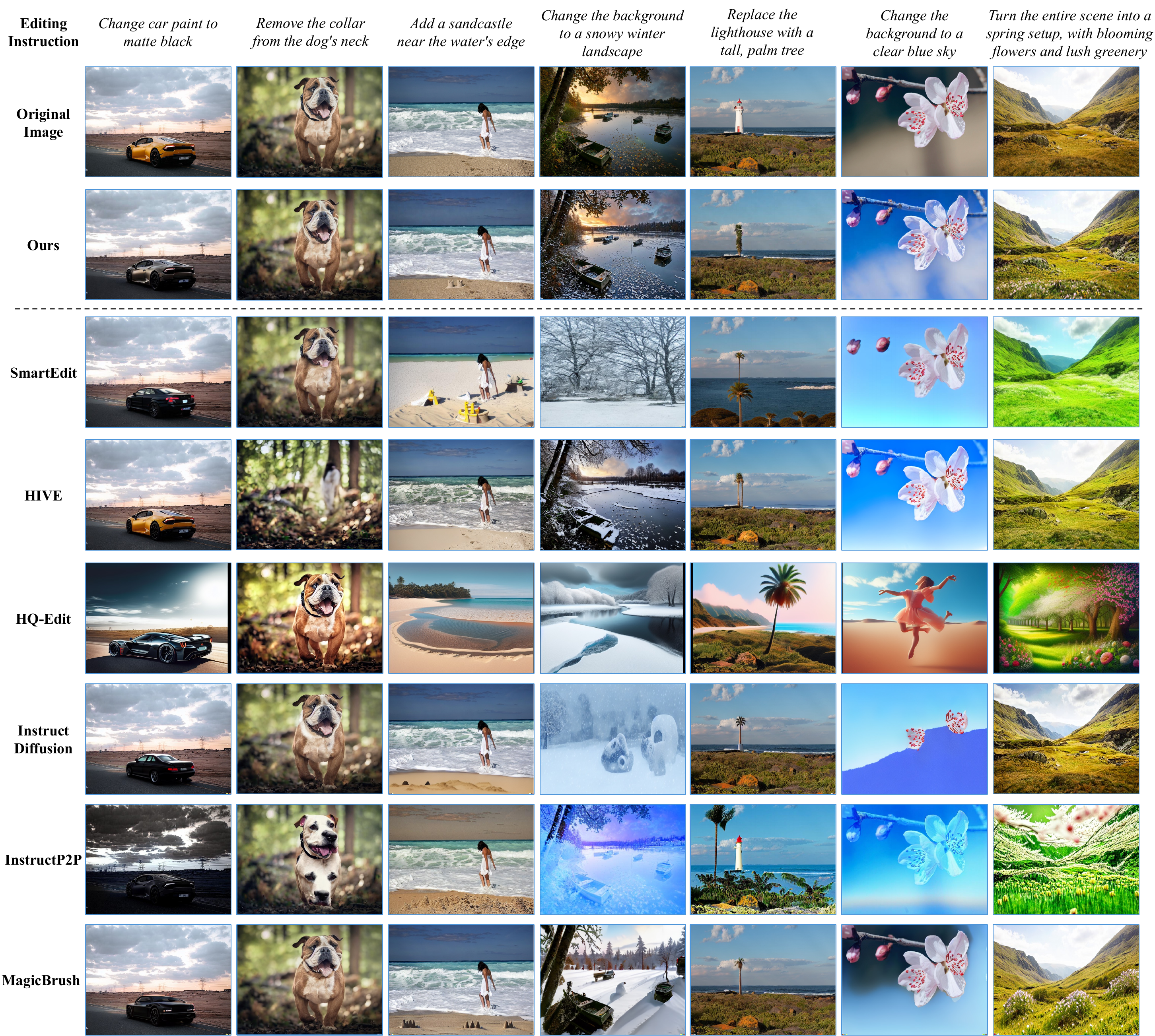}
    \caption{
        More visual comparison with existing methods.
    }
    \vspace{1.0cm}
    \label{fig:supp_visual_comparison}
\end{figure*}

\begin{figure*}[t!]\centering
    \includegraphics[width=\textwidth]{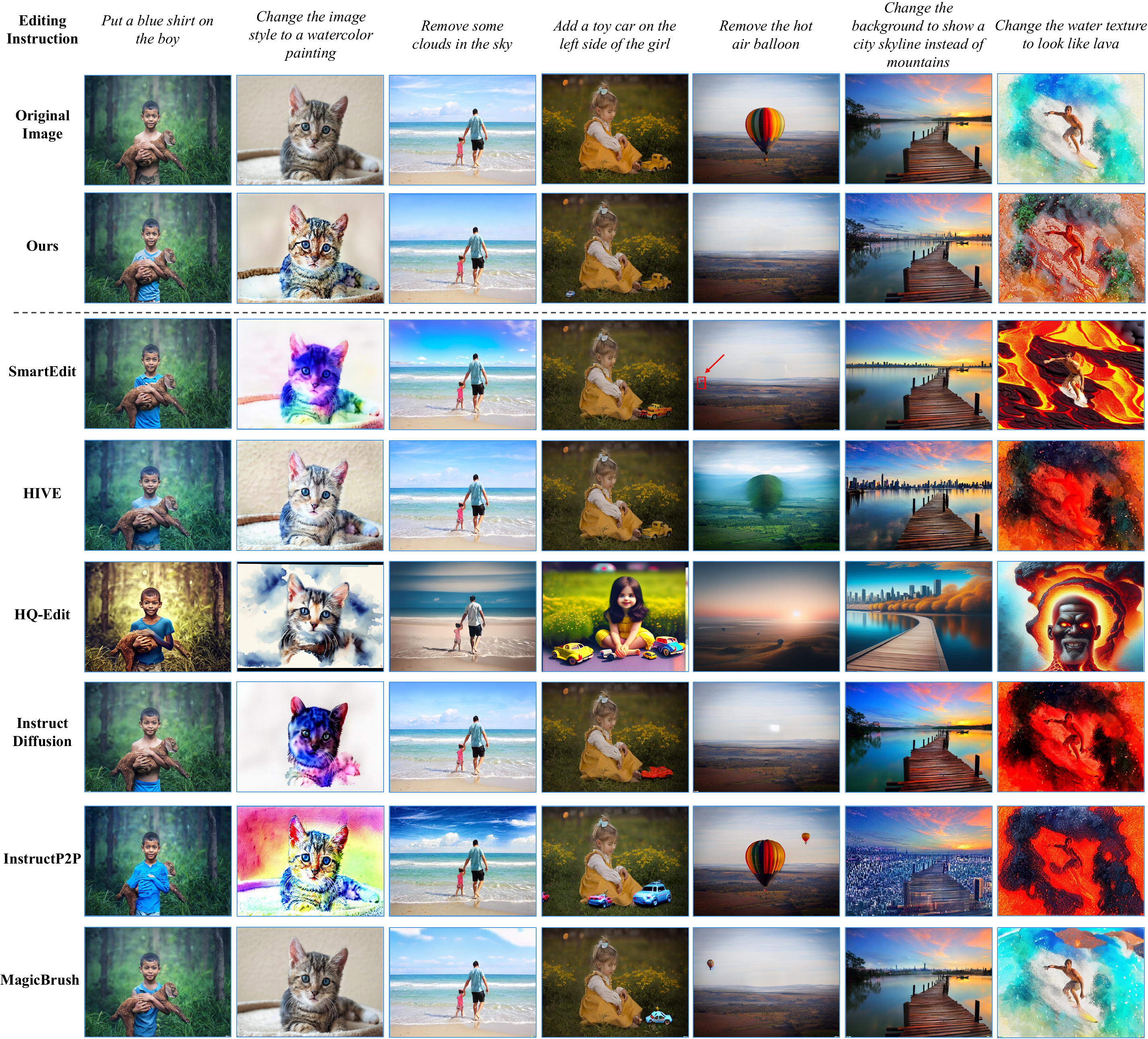}
    \caption{
        More visual comparison with existing methods.
    }
    \vspace{1.0cm}
    \label{fig:supp_visual_comparison_1}
\end{figure*}

We show more visual comparison with existing instruction-based image editing methods in Fig.~\ref{fig:supp_visual_comparison} and Fig.~\ref{fig:supp_visual_comparison_1}. Compared to existing instruction-based editing methods, our approach not only better understands and executes editing instructions but also preserves the original image’s layout and quality more effectively, thereby significantly outperforming previous methods.

\begin{figure*}[t!]\centering
    \includegraphics[width=1.0\linewidth]{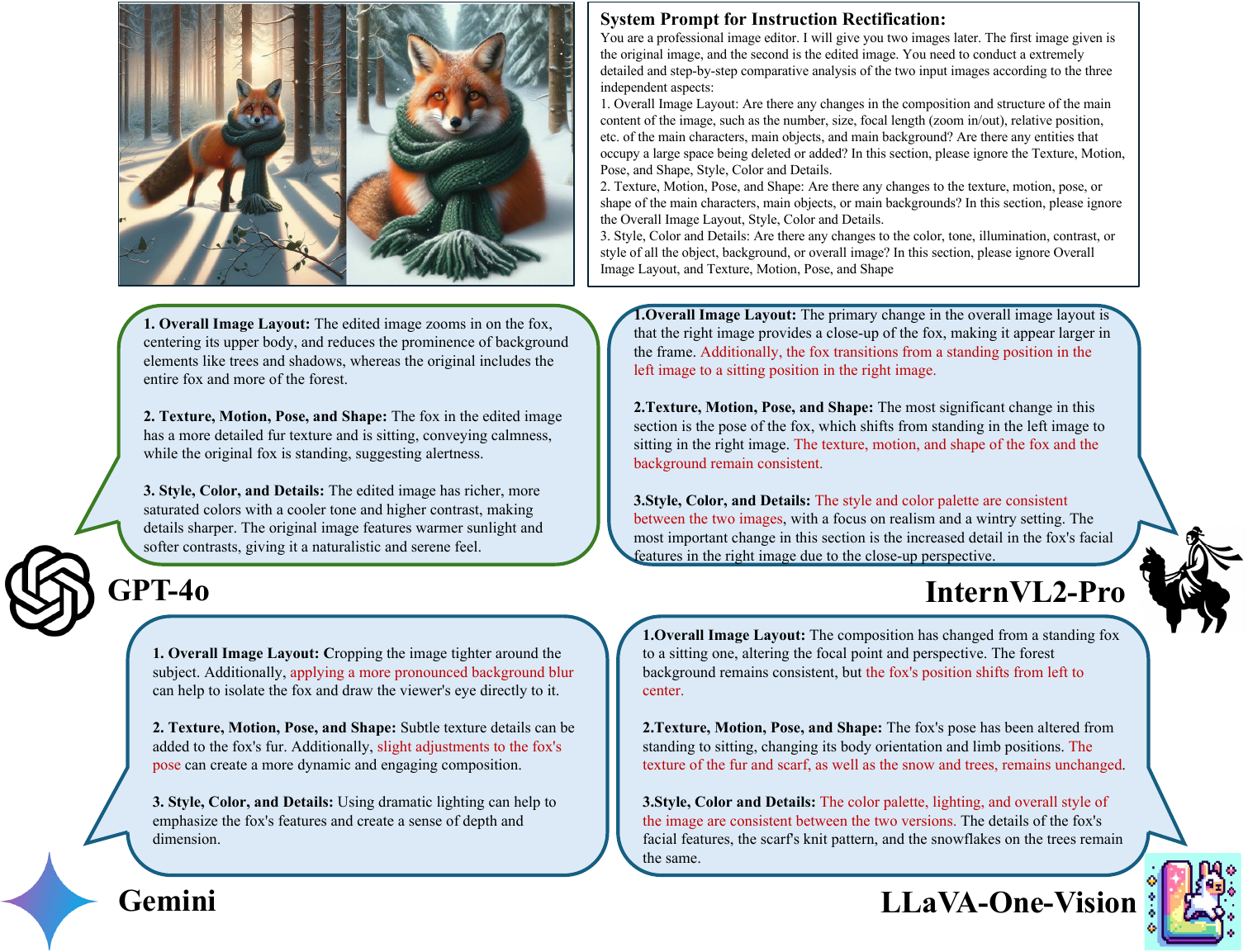}
    \vspace{-0.5cm}
    \caption{
    Comparison of different vision-language models in rectifying editing instructions based on generation prior attributes. GPT-4o achieves more stable and accurate results in describing the differences between original-edited image pairs. Text in red represents incorrectly generated instructions.
    }
    \label{fig:supp_vlms}
    \vspace{0.2cm}
    \includegraphics[width=1.0\linewidth]{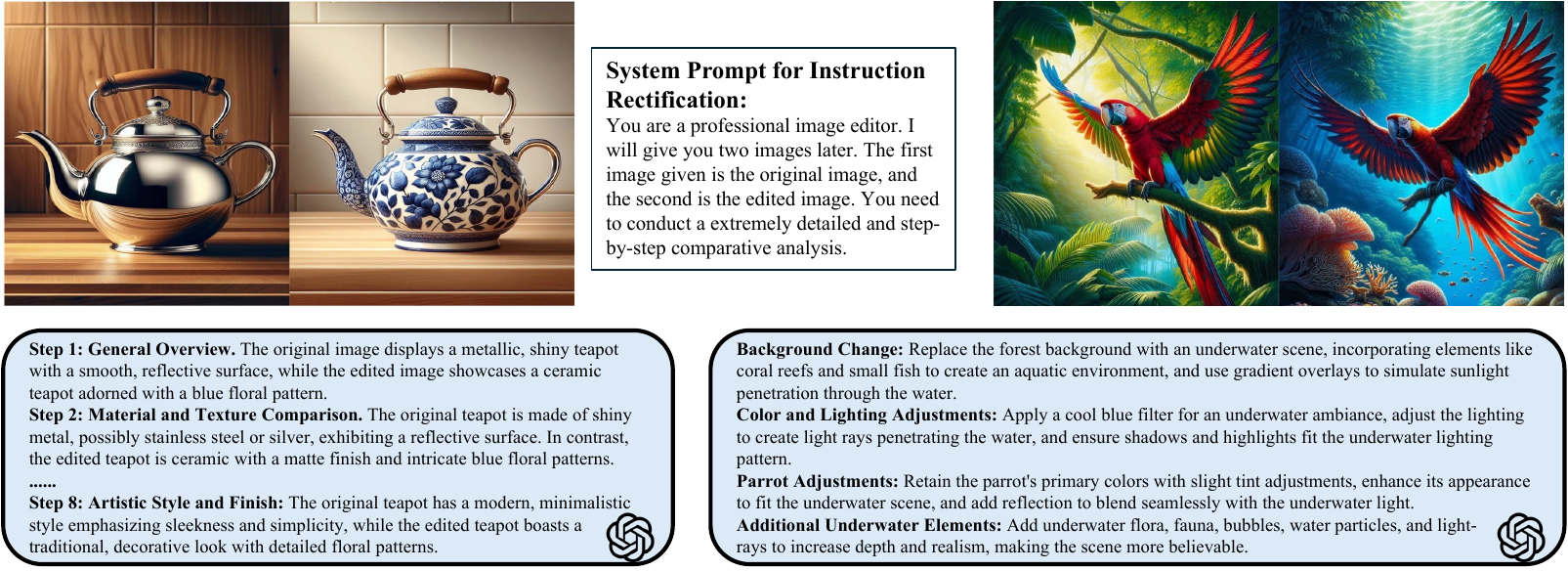}
    \vspace{-0.6cm}
    \caption{
    If the predefined four generation prior attributes are not used as templates for in-context learning, the GPT-4o rectified editing instructions will contain redundant information and lack the standardization needed for scalable processes.
    }
    \label{fig:supp_recaption_no_prior}
\end{figure*}

\vfill
\vfill

\vfill
\vfill

\end{document}